\documentclass[11pt,a4paper,english]{article}

\usepackage{wrapfig}
\usepackage{amsmath,amssymb,amsthm}
\usepackage{amsbsy}
\usepackage{bm}
\usepackage[dvipdfmx]{graphicx}
\usepackage{color}
\usepackage[scriptsize]{subfigure}

\usepackage[round]{natbib}

\newtheorem{Theorem}{Theorem}

\newtheorem{Proposition}{Proposition}

\theoremstyle{remark}
\newtheorem{Remark}{Remark}

\newcommand{\gammam}[1]{\gamma^{#1}}
\newcommand{\gammab}[1]{\gamma^{#1}}

\makeatletter
\newcommand{\figcaption}[1]{\def\@captype{figure}\caption{#1}}
\newcommand{\tblcaption}[1]{\def\@captype{table}\caption{#1}}
\makeatother

\newcommand{\Real}{{\mathbb{R}}}
\newcommand{\Realp}{\mathbb{R}_{+}}

\newcommand{\minimize}{\mathop{\mathrm{minimize}}}
\newcommand{\maximize}{\mathop{\mathrm{maximize}}}
\newcommand{\subjectto}{\textrm{subject to}}

\newcommand{\Eqref}[1]{Eq.~{\eqref{#1}}}
\newcommand{\Eqsref}[1]{Eqs.~{\eqref{#1}}}
\newcommand{\Secref}[1]{Section~{\ref{#1}}}
\newcommand{\Figref}[1]{Fig.~{\ref{#1}}}

\newcommand{\Kall}{\bar{K}}

\newcommand{\T}{^\top}

\newcommand{\prox}{{\rm prox}}

\newcommand{\propalgo}{SpicyMKL\,}

\newcommand{\ip}[2]{\left\langle #1,#2 \right\rangle}
\allowdisplaybreaks
\newcommand{\calH}{{\mathcal H}}
\newcommand{\calX}{{\mathcal X}}
\newcommand{\calY}{{\mathcal Y}}

\newcommand{\argmin}{\mathop{\mathrm{argmin}}}

\newcommand{\dd}{\mathrm{d}}

\newcommand{\boldone}{{\boldsymbol{1}}}

\newcommand{\phiCm}[1]{\phi_{C}^{(#1)}}
\newcommand{\phiC}{\phi_{C}}

\begin{document}

\title{SpicyMKL}
\author{
Taiji SUZUKI and Ryota TOMIOKA\\ \\ 
        Department of Mathematical Informatics\\
        Graduate School of Information Science and Technology\\
        The University of Tokyo\\
        \texttt{\normalsize \{t-suzuki,tomioka\}@mist.i.u-tokyo.ac.jp}}
\date{}

\maketitle

\begin{abstract}
We propose a new optimization algorithm for Multiple Kernel Learning
(MKL) called
SpicyMKL, which is applicable to general convex loss functions and
general types of regularization. 
The proposed SpicyMKL iteratively solves smooth minimization
problems. Thus,
there is no need of solving SVM, LP, or QP internally. SpicyMKL can be
viewed as a proximal
minimization method and converges super-linearly. The cost of inner
minimization is roughly
proportional to the number of active kernels. Therefore, when we aim
for a sparse kernel
combination, our algorithm scales well against increasing number of
kernels. Moreover, we
give a general block-norm formulation of MKL that includes non-sparse
regularizations, such
as elastic-net and $\ell_p$ -norm regularizations. Extending SpicyMKL, we
propose an efficient optimization method for the general regularization framework.
Experimental results show that
our algorithm is faster than existing methods especially when the
number of kernels is large
($>$ 1000).

\end{abstract}

\section{Introduction}
Kernel methods are powerful nonparametric methods in machine learning
and data analysis. Typically a kernel method fits a decision function
that lies in some Reproducing Kernel Hilbert Space
(RKHS)~\citep{TAMS:Aronszajn:1950,book:Schoelkopf+Smola:2002}.  
In such a learning framework, the choice of a kernel function can
strongly influence the performance. Instead of using a fixed kernel,
{\it Multiple Kernel Learning
(MKL)}~\citep{JMLR:Lanckriet+etal:2004,ICML:Bach+etal:2004,JMLR:MicchelliPontil:2005} aims to 
find an optimal combination of multiple candidate kernels. 

More specifically, we assume that a data point $x\in\calX$ lies in
a space $\mathcal{X}$ and we are given $M$ candidate kernel functions
$k_m:\calX\times\calX\rightarrow\Real$  ($m=1,\ldots,M$). Each kernel
function corresponds to one data source. A conical combination of $k_m$
($m=1,\ldots,M$) gives the combined kernel function
$\bar{k}=\sum_{m=1}^Md_mk_m$, where $d_m$ is a nonnegative weight. 
Our goal is to find a good set of kernel weights based on some
training examples.

While the MKL framework has opened up a possibility to combine multiple
heterogeneous data sources (numerical features, texts, links) in a
principled manner,  it is also posing an optimization
challenge:
the number  of kernels can be very large.
Instead of arguing whether polynomial
kernel or Gaussian kernel fits a given problem better, we can simply put
both of them into an MKL algorithm; instead of evaluating which band-width
parameter to choose, we can simply generate kernels from all the
possible combinations of parameter values and feed them to an MKL algorithm.
See \citet{CVPR:Gehler+Nowozin:2009,arXivSparsAcc:Tomioka+Suzuki:2009}.

Various optimization algorithms have been proposed in the context of MKL.
In the pioneering work of \citet{JMLR:Lanckriet+etal:2004}, MKL was
formulated as a semi-definite programming (SDP) problem. 
\citet{ICML:Bach+etal:2004} showed that the SDP can be reduced to a
second order conic programming (SOCP) problem. However solving SDP or
SOCP via general convex optimization solver can be quite heavy
especially for large number of samples.

More recently, {\em wrapper methods} have been proposed. A wrapper
method iteratively solves a single kernel learning problem (e.g., SVM),
for a given kernel combination and then updates the kernel weights.
A nice property of this type of methods is that it can make use of
existing well-tuned solvers for SVM.
Semi-Infinite Linear Program (SILP) approach proposed by
\citet{JMLR:Sonnenburg+etal:2006} utilizes a cutting plane method for
the update of the kernel weights.  SILP often suffers from instability
of the solution sequence especially when the number of kernels is large,  
i.e. the intermediate solution oscillates around the optimal
one~\citep{JMLR:Rakotomamonjy+etal:2008}. 
SimpleMKL proposed by \citet{JMLR:Rakotomamonjy+etal:2008} performs a
reduced gradient descent on the kernel weights. Simple MKL resolves the
drawback of SILP, but still it is a first order method. 
\citet{NIPS:Xu+etal:2008} proposed a novel Level Method (which we call
LevelMKL) as an improvement of SILP and SimpleMKL. LevelMKL is rather
efficient than SimpleMKL and scales well against the number of kernels
by utilizing sparsity  but it shows unstable behavior as the algorithm
proceeds because  LevelMKL solves Linear Programming (LP) and Quadratic
Programming (QP) of increasingly large size as its iteration proceeds. 
HessianMKL proposed by \citet{NIPSWS:Chapelle:2008} replaced the
gradient descent update of SimpleMKL with  a Newton update. 
At each iteration, HessianMKL solves a QP problem with the size of the
number of kernels to obtain the Newton 
update direction. HessianMKL shows second order convergence, but
scales badly against the number of kernels  because the size of the QP
grows as the number of kernels grows.

The first contribution of this article is an efficient optimization algorithm
for MKL based on the block 1-norm formulation introduced in
\citet{BacThiJor05} (see also \citet{ICML:Bach+etal:2004}); see
\Eqref{eq:mkl-block1-1}. The block 1-norm formulation can be viewed as a
kernelized version of group lasso~\citep{JRSS:YuanLin:2006,JMLR:BachConsistency:2008}.
%
For group lasso, or
more generally sparse estimation, efficient optimization algorithms have
recently been studied intensively, boosted by the development of compressive
sensing theory~\citep{CanRomTao06}. Based on this view, we extend the
dual augmented-Lagrangian (DAL) algorithm~\citep{ISPL:Tomioka+Sugiyama:2009}
recently proposed in the context of sparse estimation to kernel-based
learning. DAL is efficient when the number of unknown variables is much
larger than the number of samples. This enables us to scale the
proposed algorithm, which we call {\em SpicyMKL},  to thousands of
kernels. 

Compared to the original DAL
algorithm~\citep{ISPL:Tomioka+Sugiyama:2009},  our presentation is based
on an application of proximal minimization framework~\citep{Roc76} to the 
primal MKL problem. We believe that the current formulation is more
transparent compared to the dual based formulation in
\citet{ISPL:Tomioka+Sugiyama:2009}, because we are not necessarily
interested in solving the dual problem.
Moreover, we present a theorem on the rate of convergence.

SpicyMKL does not need to solve SVM, LP or QP internally as previous
approaches. Instead, it minimizes a smooth function at every step. The
cost of inner minimization is proportional to the number of active
kernels. Therefore, when we aim for a sparse kernel combination, the
proposed algorithm is efficient even when thousands of candidate kernels
are used. In fact, we show numerically that 
we are able to train a classifier with 3000 kernels in less than 10 seconds. 


Learning combination of kernels, however, has recently recognized as a
more complex task than initially thought. 
\citet{Cor09} pointed out that learning
{\em convex} kernel combination with an $\ell_1$-constraint on the kernel
weights (see \Secref{sec:framework}) produces an overly sparse (many
kernel weights are zero) solution, and it is often outperformed by a
simple uniform combination of kernels; accordingly they proposed to use
an $\ell_2$-constraint instead~\citep{UAI:Cortes+etal:2009}. In order to
search for the best trade-off between the sparse $\ell_1$-MKL and the
uniform weight combination, \citet{KloBreSonLasMueZie09} proposed a
general $\ell_p$-norm constraint and
\citet{arXivSparsAcc:Tomioka+Suzuki:2009} proposed an elastic-net
regularization,  both of which smoothly connect the
1-norm MKL and uniform weight combination.

The second contribution of this paper is to extend the block-norm
formulation that allows us to view these generalized MKL models in a
unified way, and provide an efficient optimization algorithm. We note
that while this paper was under review, \citet{KloRucBar10} presented a
slightly different approach that results in a similar optimization
algorithm. However, our formulation provides clearer relationship
between the block-norm formulation and kernel-combination weights and
is more general.

This article is organized as follows.
In Section \ref{sec:framework}, we introduce the framework of block
1-norm MKL through Tikhonov regularization on the kernel weights. In
Section \ref{sec:ALMKL}, we propose an extension of DAL algorithm to
kernel-learning setting. Our formulation of DAL algorithm is based on a
primal application of the proximal minimization framework~\citep{Roc76},
which also sheds a new light on DAL algorithm itself. Furthermore, we
discuss how we can carry out the inner minimization efficiently
exploiting the sparsity of the 1-norm MKL. In Section \ref{sec:general},
we extend our framework to general class of regularizers including the
$\ell_p$-norm MKL~\citep{KloBreSonLasMueZie09} and Elastic-net 
MKL~\citep{arXivSparsAcc:Tomioka+Suzuki:2009}. We extend the proposed
SpicyMKL algorithm for the generalized formulation and also present a
simple one-step optimization procedure for some special cases that
include Elastic-net MKL and $\ell_p$-norm MKL.
In Section \ref{sec:relationExistMeth}, we discuss the relations between the existing methods and the proposed method.
In Section \ref{sed:NumericalExp}, we show the results of numerical experiments.
The experiments show that SpicyMKL is efficient for block 1-norm regularization especially when the number of kernels is large.
Moreover the one-step optimization procedure for elastic-net regularization shows quite fast convergence. In fact, it is faster than those methods with block 1-norm regularization.
Finally, we summarize our contribution in Section \ref{sec:conclusion}.
The proof of super-linear convergence of the proposed SpicyMKL is given
in Appendix \ref{sec:ProofTheorem}.

A Matlab$^{\scriptsize \textcircled{\tiny{R}}}$ implementation of SpicyMKL is available at the following URL:
{\begin{center} \texttt{http://www.simplex.t.u-tokyo.ac.jp/\~{}s-taiji/software/SpicyMKL}\end{center}}


\section{Framework of MKL} 
\label{sec:framework} 
In this section, we first consider a learning problem with fixed
kernel weights in Section~\ref{sec:fixed}. Next in Section~\ref{sec:learningkernelweight}, using Tikhonov
regularization on the kernel weights, we derive a block 1-norm
formulation of MKL, 
which can be considered as a direct extension of
group lasso in the kernel-based learning setting. 
In addition, we discuss the connection between the current block 1-norm
formulation and the {\em squared} block 1-norm formulation. In
Section~\ref{sec:representer}, we present a finite dimensional version
of the proposed formulation and prepare notations for the later
sections.

\subsection{Fixed kernel combination}
\label{sec:fixed}
We assume that we are given $N$ samples
$
(x_i,y_i)_{i=1}^{N}
$
where $x_i$ belongs to an input space $\calX$ and $y_i$ belongs to an
output space $\calY$ (usual settings are $\calY = \{\pm 1\}$ for
classification and $\calY = \Real$ for regression).
We define the Gram matrix with respect to the kernel function $k_m$ as $K_m = (k_m(x_i,x_j))_{i,j}$. 
We assume that the Gram matrix $K_m$ is positive definite\footnote{To avoid numerical instability, 
we added  $10^{-8}$ to diagonal elements of $K_m$ in the numerical
experiments.}.

We first consider a learning problem with fixed kernel weights. More
specifically, we fix non-negative kernel weights $d_1,d_2,\ldots,d_M$ 
and consider the RKHS $\bar{\calH}$ corresponding to the
combined kernel function $\bar{k}=\sum_{m=1}^Md_mk_m$. The (squared)
RKHS norm of a function $\bar{f}$ in $\bar{\calH}$ is written as follows 
(see Sec 6 in \citet{TAMS:Aronszajn:1950}, and also \citet{JMLR:MicchelliPontil:2005}): 
\begin{align}
\label{eq:mkl-norm}
 \|\bar{f}\|_{\bar{\calH}}^2:=\min_{f_1\in\calH_1,\ldots,f_M\in\calH_M}\sum_{m=1}^{M}\frac{\|f_m\|_{\calH_m}^2}{d_m}\quad
 {\rm s.t.}\quad \bar{f}=\sum_{m=1}^Mf_m,
\end{align}
where $\calH_m$ is the RKHS that corresponds to the kernel
function $k_m$.
Accordingly, with a fixed kernel combination, a supervised learning
problem can be written as follows:
\begin{align}
\label{eq:fixed-kernel-learning}
 \minimize_{f_1\in\calH_1,\ldots, f_M\in\calH_M,b\in\Real}&\sum_{i=1}^N\ell\left(y_i,\textstyle\sum_{m=1}^Mf_m(x_i)+b\right)+\frac{C}{2}\sum_{m=1}^M\frac{\|f_m\|_{\calH_m}^2}{d_m},
\end{align}
where $b$ is a bias term and $\ell(y,f)$ is a loss function, which 
 can be the {\it hinge loss} $\max(1-yf,0)$ or the {\it logistic loss}
 $\log(1+\exp(-yf))$  for a classification problem,  or the {\it squared
 loss} $(y-f)^2$ or the {\it SVR loss} $\max( |y-f| - \epsilon, 0)$ for
 a regression problem.  
 The above formulation may not be so useful in practice, because we can
compute the combined kernel function $\bar{k}$ and optimize over
$\bar{f}$ instead of optimizing $M$ functions $f_1,\ldots,f_M$. 
However, explicitly handling the kernel weights allows us to consider various generalizations
of MKL in a unified manner.

\subsection{Learning kernel weights}
\label{sec:learningkernelweight}
In order to learn the kernel weights $d_m$ through the
objective~\eqref{eq:fixed-kernel-learning}, there is clearly a need for
regularization, because the objective is a decreasing function of the
kernel weights $d_m$. Roughly speaking, the kernel weight $d_m$
corresponds to the complexity allowed for the $m$th classifier component
$f_m$, without regularization, we can get a serious over-fitting.

One way to prevent such overfitting is to penalize the increase of the
kernel weight $d_m$ by adding a penalty term. Adding a linear penalty
term, we have the following optimization problem.
\begin{align}
\label{eq:mkl-tikhonov-l1}
  \minimize_{
\substack{
f_1\in\calH_1,\ldots,f_M\in\calH_M,\\
b\in\Real,\\
d_1\geq 0,\ldots,d_M\geq 0}}&\sum_{i=1}^N\ell\left(y_i,\textstyle\sum_{m=1}^Mf_m(x_i)+b\right)+\frac{C}{2}\sum_{m=1}^M\left(\frac{\|f_m\|_{\calH_m}^2}{d_m}+ d_m\right).
\end{align}
The above formulation reduces to the block 1-norm introduced in
\citet{BacThiJor05} by explicitly minimizing over $d_m$ as follows:
\begin{align}
\label{eq:mkl-block1-1}
  \minimize_{\substack{
f_1\in\calH_1,\ldots, f_M\in\calH_M,\\
b\in\Real}}&\sum_{i=1}^N\ell\left(y_i,\textstyle\sum_{m=1}^Mf_m(x_i)+b\right)+C\sum_{m=1}^M\|f_m\|_{\calH_m},
\end{align}
where we used the inequality of arithmetic and geometric means; the minimum with respect to $d_m$ is obtained by taking $d_m=\|f_m\|_{\calH_m}$.

The regularization term in the above block 1-norm formulation is the
linear sum of RKHS norms. This formulation can be seen as a direct
generalization of group lasso~\citep{JRSS:YuanLin:2006} to the
kernel-based learning setting, and motivates us to extend an efficient
algorithm for sparse estimation to MKL.

The block 1-norm formulation~\eqref{eq:mkl-block1-1} is related to the
following squared block 1-norm formulation considered in 
\citet{ICML:Bach+etal:2004,JMLR:Sonnenburg+etal:2006,ZieOng07,JMLR:Rakotomamonjy+etal:2008}:
\begin{align}
 \label{eq:mkl-block1-2}
  \minimize_{f_1\in\calH_1,\ldots,
 f_M\in\calH_M,b\in\Real}&\sum_{i=1}^N\ell\left(y_i,\textstyle\sum_{m=1}^Mf_m(x_i)+b\right)+\frac{\tilde{C}}{2}\left(\sum_{m=1}^M\|f_m\|_{\calH_m}\right)^2,
\end{align}
which is obtained by considering a simplex constraint on the kernel
weights~\citep{KloBreSonLasMueZie09} instead of penalizing them as in
\eqref{eq:mkl-tikhonov-l1}.

 The solution of the two problems
\eqref{eq:mkl-block1-1} and \eqref{eq:mkl-block1-2} can be mapped to
each other. In fact, let $\{f_m^\star\}_{m=1}^M$ be the minimizer of the
block 1-norm formulation~\eqref{eq:mkl-block1-1} with the regularization
parameter $C$ and let $\tilde{C}$ be 
\begin{equation}
\label{CCdashCorrespondence}
\textstyle
\tilde{C} = C(\sum_{m=1}^M \|f_m^\star \|_{\calH_m} ).
\end{equation} 
Then $\{f_m^{\star}\}$ also minimizes the squared block 1-norm
formulation~\eqref{eq:mkl-block1-2} with the regularization parameter
$\tilde{C}$ because of the relation 
$$
\partial_{f_m} \frac{1}{2}\left(\sum_{m=1}^M \|f_m \|_{\calH_m} \right)^2 =  \left(\sum_{m=1}^M \|f_m \|_{\calH_m} \right) \partial_{f_m} \|f_m \|_{\calH_m},
$$
where $\partial_{f_m}$ is a subdifferential with respect to $f_m$.


\subsection{Representer theorem}
\label{sec:representer}
In this subsection, we convert the block 1-norm MKL
formulation~\eqref{eq:mkl-block1-1} into a finite dimensional
optimization problem  via the representer theorem and prepare notation
for later sections. 

The optimal solution of \eqref{eq:mkl-block1-1} is attained 
in the form of $f_m(x) = \sum_{i=1}^N k_m(x,x_i) \alpha_{m,i}$
due to the representer theorem~\citep{JMAA:KimeldorfWahba:1971}. 
Thus, the optimization problem \eqref{eq:mkl-block1-1} is reduced to 
the following finite dimensional optimization problem:
\begin{align*}
\minimize\limits_{\alpha\in\Real^{NM},b\in\Real}\quad \sum\limits_{i=1}^N\ell(y_i, \textstyle\sum_{m=1}^M\sum_{j=1}^Nk_m(x_i,x_j)\alpha_{m,j} + b) + C \sum\limits_{m=1}^M \|\alpha_m\|_{K_m}
\end{align*}
where 
$\alpha_m = (\alpha_{m,1},\dots, \alpha_{m,N})^\top$, $\alpha =
(\alpha_1^\top,\dots,\alpha_M^\top)^{\top}\in\Real^{NM}$, and the norm
$\|\cdot\|_{K_m}$ is defined through the inner product
$\langle \alpha_m, \beta_m \rangle_{K_m} := \alpha_m^\top K_m \beta_m$
for $\alpha_m, \beta_m \in \Real^N$. We also define the norm $\|\cdot\|_{\Kall}$
through the inner product $\langle \alpha, \beta
\rangle_{\Kall}:=\sum_{m=1}^M\langle \alpha_m, \beta_m \rangle_{K_m}$ for $\alpha,\beta\in\Real^{NM}$.


For simplicity we rewrite the above problem as 
\begin{flalign}
\label{eq:MKL_primal}
\hspace{-1.3cm}
   \minimize_{\alpha\in\Real^{NM},b\in\Real} 
\quad\underbrace{L(\Kall\alpha + b\mathbf{1}) + \phiC(\alpha)}_{=:L_C(\alpha,b)},
\end{flalign}
where $\Kall =(K_1,\dots,K_M)\in\Real^{N\times NM}$, $\mathbf{1} = (1,\dots,1)^{\top}$ and  
\begin{flalign*}
&L(z) =   \sum_{i=1}^N \ell(y_i, z_i),~~~~~\\
&\phi_{C}(\alpha) =\sum\limits_{m=1}^M\phiCm{m}(\alpha_m)= C\sum\limits_{m=1}^M \|\alpha_m\|_{K_m}.
\end{flalign*}


\section{A dual augmented-Lagrangian method for MKL}
\label{sec:ALMKL}
In this section, we first present an extension of dual augmented-Lagrangian
(DAL) algorithm to kernel-learning problem through a new
approach based on the proximal minimization framework~\citep{Roc76}.
Second, assuming that the loss function is twice differentiable, we
discuss how we can compute each minimization step efficiently
in Section~\ref{sec:MinALFunc}.
Finally,  the method is extended to the situation where the loss
function is not differentiable in Section~\ref{sec:MinALfunctionGeneral}.

\subsection{MKL optimization via proximal minimization}
Starting from some initial solution $(\alpha^{(1)},b^{(1)})$, 
the proximal minimization algorithm~\citep{Roc76} iteratively minimizes
the objective \eqref{eq:MKL_primal} together with proximity terms as follows:
\begin{align}
\label{eq:prox_min}
& (\alpha^{(t+1)}\!,b^{(t+1)})\!
=\!\!\!\!   
\argmin_{\alpha\in\Real^{MN},b\in\Real}
\!\!
\Biggl(
 L_C(\alpha,b) 
+\!\! \frac{1}{2\gammab{(t)}} \Bigl( \|\alpha-\alpha^{(t)}\|_{\Kall}^2
+(b-b^{(t)})^2\Bigr)\Biggr),
\end{align}
where $0< \gamma^{(1)} \leq \gamma^{(2)} \leq \gamma^{(3)} \leq \ldots$
is a nondecreasing
sequence of {\em proximity parameters}\footnote{
Typically we exponentially increase $\gamma^{(t)}$, e.g. $\gamma^{(t)} = 2^{t}$.
In practice, we can use different values of proximity parameters for each variable (e.g. use $\gamma_m^{(t)}$ for $\alpha_m$
and $\gamma_b^{(t)}$ for $b$) and choose $\gamma$ adaptively depending on the scales of the variables.}  
and $(\alpha^{(t)},b^{(t)})$ is an approximate minimizer at the $t$th
iteration; $L_C$ is the (regularized) objective
function~\eqref{eq:MKL_primal}. 
The last two terms in the right-hand side are {\em proximity terms} that
tries to keep the next solution $(\alpha^{(t+1)},b^{(t+1)})$ close to the
current solution $(\alpha^{(t)},b^{(t)})$. Thus, we call the minimization problem~\eqref{eq:prox_min}
a proximal MKL problem. 
Solving the proximal MKL problem~\eqref{eq:prox_min} seems as difficult as solving the original MKL problem in the primal.
However, when we consider the dual
problems, solving the dual of proximal MKL problem~\eqref{eq:prox_min} is a {\em smooth}
minimization problem and can be minimized efficiently, whereas solving
the dual of the original MKL problem~\eqref{eq:MKL_primal} is a non-smooth minimization
problem and not necessarily easier than solving 
the primal. 

The update equation can be interpreted as an {\em implicit} gradient
method on the objective function $L_C(\alpha,b)$. In fact, by taking the
subgradient of the update equation~\eqref{eq:prox_min} and equating it to
zero, we have
\begin{align*}
\alpha_m^{(t+1)} &\in \alpha_m^{(t)} - \gammam{(t)} K_m^{-1}\partial_{\alpha_{m}} L_C(\alpha^{(t+1)},b^{(t+1)}), \\ 
b^{(t+1)} &\in b^{(t)} - \gammab{(t)} \partial_{b} L_C (\alpha^{(t+1)},b^{(t+1)}).
\end{align*}
This implies the $t$th update step is
a subgradient  of the original objective function $L_C$ at the {\it next} solution $(\alpha_m^{(t+1)},b^{(t+1)})$.

The super-linear convergence of proximal minimization algorithm~\citep{Roc76,Book:Bertsekas:1982,JMLR:Tomioka+Suzuki+Sugiyama:2011}
can also be extended to kernel-based learning setting as in the
following theorem.
\begin{Theorem}
\label{th:superlinear}
Suppose the problem \eqref{eq:MKL_primal} has a unique\footnote{
The uniqueness of the optimal solution is just for simplicity. 
The result can be generalized for a compact optimal solution set (see \citep{Book:Bertsekas:1982}).} 
optimal solution $\alpha^{\ast}$, $b^{\ast}$,
and 
there exist a scalar $\sigma > 0 $ and a $\delta$-neighborhood ($\delta > 0$) of the optimal solution such that 
\begin{equation}
\label{eq:LocalStrongConv}
L_C(\alpha,b)\geq
L_C(\alpha^{\ast},b^{\ast})
+ 
\sigma( \|\alpha - \alpha^{\ast} \|_{\Kall}^2 + (b - b^{\ast})^2 ),
\end{equation}
for all $(\alpha,b) \in \Real^{MN} \times \Real$ satisfying $\|\alpha - \alpha^{\ast} \|_{\Kall}^2 + (b - b^{\ast})^2  \leq \delta^2$.
Then for all sufficiently large $t$ we have 
\[
\frac{\|\alpha^{(t+1)} - \alpha^{\ast} \|_{\Kall}^2 + (b^{(t+1)} - b^{\ast})^2 }{ \|\alpha^{(t)} - \alpha^{\ast} \|_{\Kall}^2 + (b^{(t)} - b^{\ast})^2}
\leq \frac{1}{(1+ \sigma \gamma^{(t)})^2}. 
\]
Therefore if $\displaystyle \lim_{t\to \infty}\gamma^{(t)} = \infty$, the solution $(\alpha^{(t)},b^{(t)})$ converges to the optimal solution super-linearly. 
\end{Theorem}
\begin{proof}
The proof is given in Appendix \ref{sec:ProofTheorem}.
\end{proof}

\subsection{Derivation of SpicyMKL}
\label{sec:dal-prox}
\begin{figure}[tb]
\begin{center}
\includegraphics[width=5cm,clip]{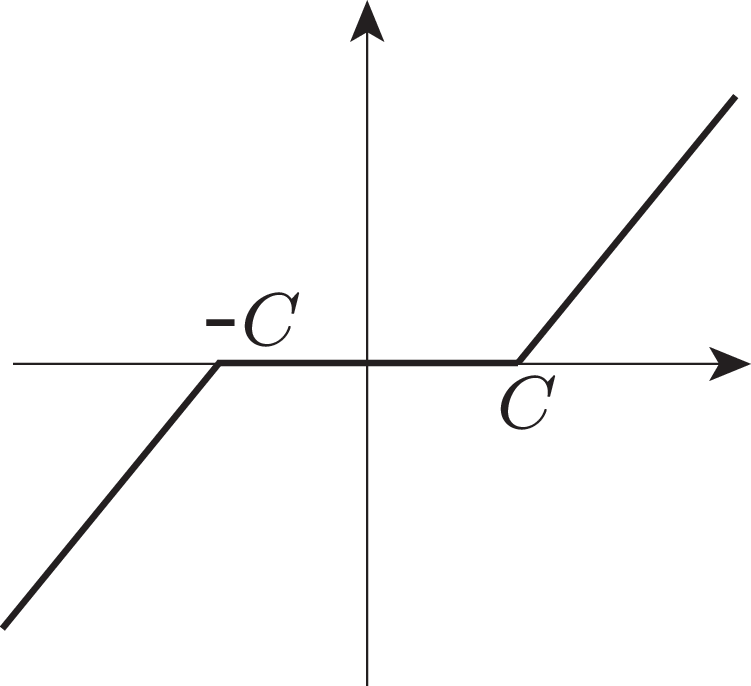}
\caption{Illustration of soft thresholding function $\prox(\cdot|\phi_{C}^{(m)})$.}
\label{fig:ST}
\end{center}
\end{figure}

Although directly minimizing the proximal MKL problem~\eqref{eq:prox_min}
is not a trivial task, its dual problem can efficiently be solved.
Once we solve the dual of the proximal minimization
update~\eqref{eq:prox_min}, we can update the primal variables
$(\alpha^{(t)},b^{(t)})$. The resulting iteration can be written as 
follows:
\begin{align}
\label{eq:update_rho}
 \rho^{(t)}&:=\argmin_{\rho\in\Real^{N}}
\varphi_{\gammam{(t)}}(\rho;\alpha^{(t)},b^{(t)}),\\
\label{eq:update_alpha}
\alpha_m^{(t+1)}&=\prox(\alpha_m^{(t)}+\gammam{(t)}\rho^{(t)}|\phi_{\gamma^{(t)}C}^{(m)})\qquad(m=1,\ldots,M),\\
\label{eq:update_b}
b^{(t+1)}&=b^{(t)}+\gammab{(t)}\sum_{i=1}^N\rho_i^{(t)},
\end{align}
where $\varphi_{\gamma}$ is the dual objective, which we derive in the
sequel, and the proximity operator $\prox(\cdot|\phi_{C}^{(m)})$
corresponding to the regularizer $\phi_{C}^{(m)}$  is defined as follows:
\begin{align}
 \prox(v_m|\phi_{C}^{(m)})&=\argmin_{v_m'\in\Real^N}\left(\phi_{C}^{(m)}(v_m')+\frac{1}{2}\|v_m'-v_m\|_{K_m}^2\right)\nonumber\\
\label{eq:softth}
&=
\begin{cases}
 0          & (\textrm{if $\|v_m\|_{K_m}\leq C$}),\\
 \frac{\|v_m\|_{K_m}-C}{\|v_m\|_{K_m}}v_m & (\textrm{otherwise}).
\end{cases}
\end{align}
The above operation is known as the soft-thresholding function in the
field of sparse
estimation (see~\citet{FigNow03,DauDefMol04}) and it has been applied to
MKL earlier in \citet{NIPSWS:Mosci+etal:2008}. Intuitively speaking, it
thresholds a vector smaller than $C$ in norm to zero but also shrinks a
vector whose norm is larger than $C$ so that the transition at
$\|v_m\|_{K_m}=C$ is continuous (soft); see Figure \ref{fig:ST} for a one
dimensional illustration.  

At every iteration we minimize the inner objective $\varphi_\gamma$ (the
dual of the proximal MKL problem~\eqref{eq:prox_min}), and use
the minimizer $\rho^{(t)}$ to update the primal variables
$(\alpha^{(t)},b^{(t)})$. The overall algorithm is shown in
Table~\ref{tab:Algorithm1}. 

\begin{table*}[tb]
\caption{Algorithm of \propalgo for block 1-norm MKL}
\label{tab:Algorithm1}
\begin{center}
\framebox{
  \begin{minipage}[t]{0.45\textwidth}
    \begin{tabbing}
      XX\=\kill
      1.\>Choose a sequence $\gammam{(t)}\to \infty$ as $t\to \infty$.\\ 
      2.\>Minimize the augmented Lagrangian with respect to $\rho$:    \\
      \phantom{2.} $\rho^{(t)} \!= \! \mathop{\argmin}_{\rho}\left( \! L^{\ast}(-\rho) \!+\! \!
\frac{1}{2\gammam{(t)}} \sum_m \| \prox(\alpha_m^{(t)}+\gammam{(t)}\rho^{(t)}|\phi_{\gammam{(t)}C}^{(m)}) \|_{K_m}^2
 \!\! + \frac{1}{2\gammab{(t)}} (b^{(t)}+\gammab{(t)}\sum_i \rho_i )^2\right).$\\
      3.\>Update $\alpha_m^{(t+1)} \leftarrow \prox(\alpha_m^{(t)}+\gammam{(t)}\rho^{(t)}|\phi_{\gammam{(t)}C}^{(m)}),~
b^{(t+1)}  \leftarrow b^{(t)} + \gammab{(t)} \sum_i \rho_i^{(t)}.$ \\
      4.\>Repeat 2.~and 3.~until the stopping criterion is satisfied. 
    \end{tabbing}
 \end{minipage}
}
\end{center}
\end{table*} 
\paragraph{Quick overview of the derivation of the iteration~\eqref{eq:update_rho}-\eqref{eq:update_b}.}
Consider the following Lagrangian of the proximal MKL
problem~\eqref{eq:prox_min}:
\begin{align}
\label{eq:Lagrangian}
 \mathcal{L}&=L(z)+\phi_{C}(\alpha)+\frac{\|\alpha-\alpha^{(t)}\|_{\Kall}^2}{2\gamma^{(t)}}+\frac{(b-b^{(t)})^2}{2\gamma^{(t)}}+\rho\T(z-\Kall\alpha-b\mathbf{1}),
\end{align}
where $\rho\in\Real^{N}$ is the Lagrangian multiplier corresponding to
the equality constraint $z=\bar{K}\alpha+b\mathbf{1}$. The vector $\rho^{(t)}$
in the first step~\eqref{eq:update_rho} is the optimal Lagrangian
multiplier that {\em maximizes} the dual of the proximal MKL
problem~\eqref{eq:prox_min} (see \Eqref{eq:alfunc}). The remaining
steps~\eqref{eq:update_alpha}--\eqref{eq:update_b} can be obtained by
minimizing the above Lagrangian with respect to $\alpha$ and $b$ for
$\rho=\rho^{(t)}$, respectively.

\paragraph{Detailed derivation of the
    iteration~\eqref{eq:update_rho}-\eqref{eq:update_b}.}
Let's consider the constraint reformulation of the proximal MKL
problem~\eqref{eq:prox_min} as follows:
\begin{align*}
 \minimize_{
\begin{subarray}{l}
\alpha\in\Real^{MN},b\in\Real\\
z\in\Real^{N}
\end{subarray}}
\quad &L(z)+\phi_{C}(\alpha)+\frac{\|\alpha-\alpha^{(t)}\|_{\Kall}^2}{2\gamma^{(t)}}+\frac{(b-b^{(t)})^2}{2\gamma^{(t)}},\\
\subjectto \quad&z=\Kall\alpha+b\mathbf{1}.
\end{align*}
The Lagrangian of the above constrained minimization problem can be
written as in \Eqref{eq:Lagrangian}.

The dual problem can be derived by minimizing the
Lagrangian~\eqref{eq:Lagrangian} with respect to the primal variables
$(z,\alpha, b)$. See \Eqref{eq:alfunc} for the final expression. Note that the minimization is separable into
minimization with respect to $z$ (\Eqref{eq:derivedual1}), $\alpha$
(\Eqref{eq:derivedual2}), and $b$ (\Eqref{eq:derivedual3}).

First, minimizing the Lagrangian with respect to $z$ gives
\begin{align}
\label{eq:derivedual1}
 \min_{z\in\Real^N}\left(L(z)+\rho\T z\right)=-L^{\ast}(-\rho),
\end{align}
where $L^{\ast}$ is the convex conjugate of the loss function $L$ as
follows:
\begin{align*}
 L^\ast(-\rho)&:=\sup_{z\in\Real^N}\left((-\rho)\T z- L(z)\right)\nonumber\\
 &=\sum_{i=1}^N\left(-\rho_iz_i-\ell(y_i,z_i)\right)\nonumber\\
 &=\sum_{i=1}^N\ell^\ast(y_i,-\rho_i),\nonumber
\end{align*}
where $\ell^\ast$ is the convex conjugate of the loss $\ell$
with respect to the second argument.

For example, the conjugate loss $\ell_L^\ast$  for the logistic loss $\ell_L$ is
the negative entropy function as follows:
\begin{align}
\label{eq:loss_logit_conj}
 \ell_L^{\ast}(y,-\rho)&=
\begin{cases}
(y\rho)\log(y\rho)+(1-y\rho)\log(1-y\rho)  & (\textrm{if $0\leq
 y\rho\leq 1$}),\\
+\infty & (\textrm{otherwise}).
\end{cases}
\end{align}
The conjugate loss $\ell_H^\ast$ for the hinge loss $\ell_H$
 is given as follows:
\begin{align}
\label{eq:loss_hinge_conj}
 \ell_H^{\ast}(y,-\rho)&=
\begin{cases}
 -y\rho & (\textrm{if $0\leq y\rho\leq 1$}),\\
 +\infty & (\textrm{otherwise}).
\end{cases}
\end{align}

\begin{figure}[tb]
 \begin{center}
  \includegraphics[width=.4\textwidth]{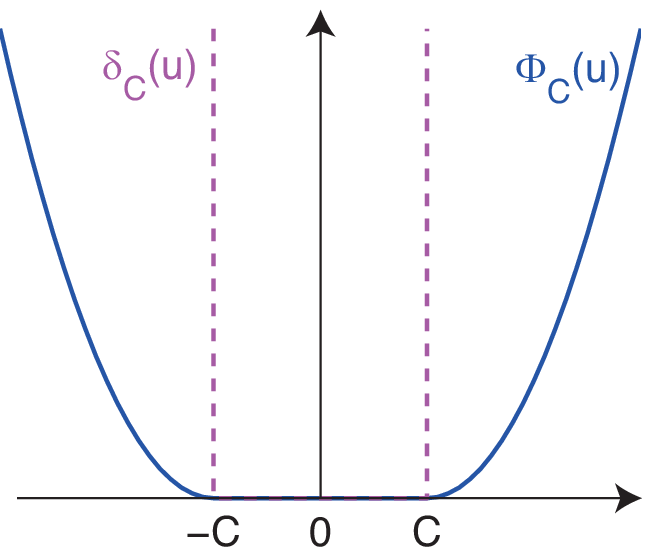}
  \caption{Comparison of the conjugate regularizer $\delta_C^{(m)}$ and the
  corresponding Moreau's envelope function $\Phi_C^{(m)}$ in one
  dimension. The conjugate regularizer $\delta_C^{(m)}$~\eqref{eq:Aphistar} is
  nondifferentiable at the boundary of its domain, whereas its envelope
  function $\Phi_C^{(m)}$ is smooth.}
  \label{fig:envelope}
 \end{center}
\end{figure}

Second, minimizing the Lagrangian~\eqref{eq:Lagrangian} with respect to
$\alpha$, we obtain
\begin{align}
&\min_{\alpha\in\Real^{MN}}\left(\phi_{C}(\alpha)+\frac{\|\alpha-\alpha^{(t)}\|_{\Kall}^2}{2\gamma^{(t)}}-\rho\T\Kall\alpha\right)\nonumber\\
\label{eq:derivedual2phi}
&=\min_{\alpha\in\Real^{MN}}\left(\phi_{C}(\alpha)+\frac{\|\alpha-\alpha^{(t)}-\gamma^{(t)}\rho\|_{\Kall}^2}{2\gamma^{(t)}}\right)-\frac{\|\alpha^{(t)}+\gamma^{(t)}\rho\|_{\Kall}^2}{2\gamma^{(t)}}+{\rm const}\\
\label{eq:derivedual2delta}
&\mathop{=}^{\eqref{eq:Aphistar},\eqref{eq:MoreauConvolution}}-\frac{1}{\gamma^{(t)}}\min_{u\in\Real^{MN}}\left(
\delta_{\gamma^{(t)}C}(u)+\frac{\|u-\alpha^{(t)}-\gamma^{(t)}\rho\|_{\Kall}^2}{2}
\right)+{\rm const}\\
\label{eq:derivedual2}
&\mathop{=}^{\eqref{eq:Phi-l1}}
-\frac{1}{\gamma^{(t)}}\sum_{m=1}^{M}\Phi_{\gamma^{(t)}C}^{(m)}(\alpha_m+\gamma^{(t)}\rho)+{\rm const.},
\end{align}
where ${\rm const}$ is a term that only depends on $\alpha^{(t)}$ and
$\gamma^{(t)}$; the function $\delta_C:\Real^{MN}\rightarrow\Real$ is the convex conjugate of the regularization term
$\phiCm{m}(\alpha_m)=C\|\alpha_m\|_{K_m}$ as follows:
\begin{align}
\delta_C(u)&:=\sum_{m=1}^{M}\delta_C^{(m)}(u_m),\nonumber
\intertext{where}
\delta_C^{(m)}(u_m)
 &:=\sup_{\alpha_m\in\Real^N}\left(\ip{u_m}{\alpha_m}_{K_m}-\phi_{C}^{(m)}(\alpha_m)\right)\nonumber\\
&=\sup_{\alpha_m\in\Real^N}\left(\|u_m\|_{K_m}\|\alpha_m\|_{K_m}-C\|\alpha_m\|_{K_m}\right)\nonumber\\
&=
\label{eq:Aphistar}
\begin{cases}
 0 & (\textrm{if $\|u_m\|_{K_m}\leq C$}),\\
+\infty & (\textrm{otherwise}).
\end{cases}
\end{align}
See \Figref{fig:envelope} for a one dimensional illustration of the
conjugate regularizer $\delta_C^{(m)}$.	In addition, the function
$\Phi_C^{(m)}$ in the last line is {\it Moreau's envelope function} (see \Figref{fig:envelope}): 
\begin{align}
\Phi_{C}^{(m)}(v_m)&:=\min\limits_{v'_m\in\Real^{N}}\left(
\delta_{C}^{(m)}(v'_m)+\frac{1}{2}\|v'_m-v_m \|_{K_m}^2
\right),\nonumber\\
\label{eq:Phi-l1}
&=
\left\|\prox(v_m |\phi_{C}^{(m)})\right\|_{K_m}^2.
\end{align}  
See \Eqref{eq:softth} and Figure~\ref{fig:ST} for the definition of the
soft-threshold operation $\prox(\cdot|\phi_{C}^{(m)})$.
Furthermore, we used the following proposition and some algebra to derive \Eqref{eq:derivedual2delta}
from \Eqref{eq:derivedual2phi}. 
\begin{Proposition}
Let $f:\Real^{n}\rightarrow \Real$ be a closed proper convex function
 and $f^{\ast}$ be the convex conjugate of $f$ defined as
\begin{align*}
 f^{\ast}(y)&=\sup_{x\in\Real^{n}}\left(\ip{y}{x}_{K}-f(x)\right),
\end{align*}
where $K\in\Real^{n\times n}$ is a positive semidefinite matrix.  Then 
\begin{align}
 \min_{x\in\Real^n}\left(f(x)+\frac{\|x-z\|_{K}^2}{2}\right)+\min_{y\in\Real^n}\left(f^{\ast}(y)+\frac{\|y-z\|_{K}^2}{2}\right)&=\frac{\|z\|_{K}^2}{2}.
\label{eq:MoreauConvolution}
\end{align}
\end{Proposition}
\begin{proof}
It is a straightforward generalization of Moreau's theorem. See
 \citet[Theorem 31.5]{Book:Rockafellar:ConvexAnalysis} and \citet{JMLR:Tomioka+Suzuki+Sugiyama:2011}.
\end{proof}

Finally, minimizing the Lagrangian~\eqref{eq:Lagrangian} with respect to
$b$, we obtain
\begin{align}
\label{eq:derivedual3}
\min_{b\in\Real}\left(\frac{(b-b^{(t)})^2}{2\gamma^{(t)}}-b\rho\T\mathbf{1}\right)&=-\frac{(b^{(t)}+\gamma^{(t)}\rho\T\mathbf{1})^2}{2\gamma^{(t)}}+{\rm
 const},
\end{align}
where const is a term that only depends on $b^{(t)}$ and $\gamma^{(t)}$.

Combining Eqs.~\eqref{eq:derivedual1}, \eqref{eq:derivedual2}, and
\eqref{eq:derivedual3}, the dual of the proximal MKL
problem~\eqref{eq:prox_min} can be obtained as follows:
\begin{align}
\label{eq:alfunc}
 \maximize_{\rho\in\Real^N}\quad&-L^{\ast}(-\rho)-\frac{1}{2\gamma^{(t)}}\sum_{m=1}^{M}\left\|\prox(\alpha_m^{(t)}+\gamma^{(t)}\rho|\phi_{\gamma C}^{(m)})\right\|_{K_m}^2
-\frac{1}{2\gamma^{(t)}}\left(b^{(t)}+\gamma^{(t)}\textstyle\sum_{i=1}^N\rho_i\right)^2,
\end{align}
where the constant terms are ignored. We denote the maximand in the above dual
problem by $-\varphi_{\gamma^{(t)}}(\rho;\alpha^{(t)},b^{(t)})$; see \Eqref{eq:update_rho}.

\subsection{Minimizing the inner objective function}
\label{sec:MinALFunc}
The inner objective function \eqref{eq:alfunc} that we
need to minimize at every iteration is convex and {\em differentiable}
when $L^{\ast}$ is differentiable. In fact, the gradient and the Hessian
of the inner objective  $\varphi_{\gamma}$ can be written as follows:
\begin{align}
\label{eq:varphi_grad}
&\nabla_{\rho} \varphi_\gamma(\rho;\alpha,b) 
=-
\nabla_{\rho}L^{\ast}(-\rho)+ (b+\gamma\sum_i \rho_i) \boldone 
+ \sum_{m\in M_+} K_m\prox\left(\alpha_m+\gamma\rho\Big|\phi_{\gamma C}^{(m)}\right), 
\\
&\nabla^2_{\rho} \varphi_\gamma(\rho;\alpha,b) 
=
\nabla^2_{\rho}L^{\ast}(-\rho) + \gamma \boldone \boldone^\top 
+ \gamma \!\!\!\!\sum_{m\in M_+} \!\!\!\! 
 \left((1-q_m)K_m\!\!+\!q_mK_m\tilde{v}_m\tilde{v}_m\T K_m\right),
 \label{eq:varphi_hess}
\end{align}
where $M_+$ is the set of indices corresponding to the {\em active
kernels}; i.e., $M_+=\{m\in \{1,\ldots,M\} \mid
\|\alpha_m+\gammam{}\rho\|_{K_m}>\gammam{} C \}$; 
for $m\in M_+$, a scalar $q_m$ and a vector $\tilde{v}_m\in\Real^N$ are
defined as $q_m:=\frac{\gammam{}C}{\|\alpha_m+\gammam{}\rho 
\|_{K_m}}$ and 
$\tilde{v}_m:=(\alpha_m+\gammam{} \rho)/\| \alpha_m+\gammam{} \rho \|_{K_m}$.

\begin{Remark}
{\rm
The computation of the objective
 $\varphi_\gamma(\rho;\alpha,b)$~\eqref{eq:alfunc}, the
 gradient~\eqref{eq:varphi_grad}, and the Hessian~\eqref{eq:varphi_hess}
 is efficient because they require only the terms corresponding to
 the {\it active kernels} $\{k_m \mid m\in M_+\}$. 
}
\end{Remark}

The above sparsity, which makes the proposed algorithm efficient, comes
from our dual formulation. 
By taking the dual, there appears {\it flat} region (see
\Figref{fig:envelope});  i.e. the region 
$\{\rho \mid \prox(\alpha_m^{(t)}+\gammam{(t)}\rho|\phi_{\gammam{(t)}C}^{(m)})=0\}$ is not a single point but has its interior. 

Since the inner objective function \eqref{eq:alfunc} is differentiable,
and the 
sparsity of the intermediate solution can be exploited to evaluate the
gradient and Hessian of the inner objective,  the inner minimization 
can efficiently be carried out.
We call the proposed algorithm {\it Sparse Iterative MKL (SpicyMKL)}. 
The update equations \eqref{eq:update_alpha}-\eqref{eq:update_b} 
exactly correspond to the augmented Lagrangian method for the dual of the problem \eqref{eq:MKL_primal} (see
\citet{ISPL:Tomioka+Sugiyama:2009}) but derived in more general way using
the techniques from \citet{Roc76}. 

We use the Newton method with line search  for the minimization of the inner
objective~\eqref{eq:alfunc}. The line search is used to keep $\rho$ inside
the domain of the dual loss $L^\ast$.
This works when the gradient of the dual loss $L^\ast$ is unbounded at
the boundary of its domain, for example the logistic
loss~\eqref{eq:loss_logit_conj}, in which case the  minimum is never attained at
the boundary. On the other hand, for the hinge loss~\eqref{eq:loss_hinge_conj}, 
the solution lies typically at the boundary of the domain $0\leq y\rho\leq 1$.
This situation is handled separately in the next subsection.

\subsection{Explicitly handling boundary constraints}
\label{sec:MinALfunctionGeneral}
The Newton method with line search described in the last section is
unsuitable when the conjugate loss function $\ell^\ast(y,\cdot)$ has a
non-differentiable point in the interior of its domain or it has finite
gradient at the boundary of its domain. We use the same augmented
Lagrangian technique for these cases. More specifically we introduce
additional primal variables so that the AL function
$\varphi_\gamma(\cdot;\alpha,b)$ becomes differentiable. 
First we explain this in the case of hinge loss for classification. 
Next we discuss a generalization to other
cases.
\subsubsection{Optimization for hinge loss}
Here we explain how the augmented Lagrangian technique is applied to the hinge loss. 
To this end, we introduce two sets of slack
variables $\xi=(\xi_1,\ldots,\xi_N)\T\geq 0, \zeta=(\zeta_1,\ldots,\zeta_N)\T\geq 0$ as in standard
SVM literature (see e.g., \citet{book:Schoelkopf+Smola:2002}). The basic update equation
(\Eqref{eq:prox_min}) is rewritten as follows\footnote{$\Realp$ is the set of non-negative real numbers}:
\begin{align*}
&(\alpha^{(t+1)},b^{(t+1)},\xi^{(t+1)},\zeta^{(t+1)})  = \!\!\!\!\!\! \\
&
\argmin\limits_{
\begin{subarray}{l}
\alpha\in\Real^{(MN)},b\in\Real\\ \xi\in\Realp^{N},\zeta\in\Realp^{N}
\end{subarray}
}
\Biggl\{H_C(\alpha,b,\xi,\zeta)+
\frac{1}{2\gammam{(t)}} \left(
 \|\alpha-\alpha^{(t)}\|_{\Kall}^2 
+(b-b^{(t)})^2 
+\|\xi-\xi^{(t)}\|^2+\|\zeta-\zeta^{(t)}\|^2\right)\Biggr\},
\end{align*}
where
\begin{align*}
&H_C(\alpha,b,\xi,\zeta)=
\begin{cases}
\sum_{i=1}^N\xi_i
+\phiC(\alpha)
&\Big(\textrm{if } y_i(\displaystyle(\sum_{m=1}^MK_m\alpha_m)_i+b)=1-\xi_i+\zeta_i,~\xi_i \geq 0,~\zeta_i \geq 0, \forall i\Big),\\
\textstyle
+\infty&\textrm{(otherwise).}
\end{cases}
\intertext{This function $H_C$ can again be 
expressed in terms of maximum over $\rho\in\Real^N$, $u \in \Real^{MN}$ as follows:}
&H_C(\alpha,b,\xi,\zeta) \\
&= \max_{\rho\in\Real^N,u \in \Real^{MN} }\Biggl\{
-\sum_{i=1}^N(-y_i\rho_i) -b\sum_{i=1}^N\rho_i  -\sum_{m=1}^M \delta_C^{m}(u_m)
-\sum_{m=1}^M\langle \alpha_m, \rho-u_m \rangle_{K_m}
\\
&
~~~~~~~~~~~~~~~~~~~~~~+\sum_{i=1}^N\xi_i(1-y_i\rho_i) + \sum_{i=1}^N\zeta_i(y_i\rho_i)
\Biggr\}.
\end{align*}
We exchange the order of minimization and maximization as before 
and remove $\alpha,b,\xi,\zeta$, and $u_m$ by explicitly
 minimizing or maximizing over them
 (see also~\Secref{sec:dal-prox}). 
Finally we obtain the following
 update equations.
\begin{align}
\textstyle
 \alpha_m^{(t+1)}&
\textstyle
=\prox(\alpha_m^{(t)} + \gammam{(t)}\rho^{(t)} | \phi_{\gamma^{(t)}C}^{(m)}), 
\label{eq:update_hinge_one}
\\
\textstyle
b^{(t+1)} 
&=b^{(t)}+\gammab{(t)}\sum_{i=1}^N\rho_i^{(t)},
\\
\textstyle
\xi_i^{(t+1)} 
&\textstyle
=\max(0,\xi_i^{(t)}-\gamma_\xi^{(t)}(1-y_i\rho_i^{(t)})), \\
~~~~ 
\textstyle
\zeta_i^{(t+1)}
\textstyle
&=\max(0,\zeta_i^{(t)}-\gamma_\zeta^{(t)}y_i\rho_i^{(t)}).
\label{eq:update_hinge_two}
\end{align}
and 
$\rho^{(t)}\in\Real^N$ is the minimizer of the
function
 $\varphi_{\gamma^{(t)}}(\rho;\alpha^{(t)},b^{(t)},\xi^{(t)},\zeta^{(t)})$
 defined as follows:
\begin{align} 
\textstyle
\varphi_{\gamma}(\rho;\alpha,b,\xi,\zeta) = & \textstyle
 -\sum\limits_{i=1}^Ny_i\rho_i 
 +\frac{1}{2\gammab{}}(b+\gammab{}\sum\limits_{i=1}^N\rho_i)^2 
+ \sum\limits_{m=1}^M\frac{1}{2\gammam{}} \|\prox(\alpha_m + \gammam{}\rho | \phi_{\gamma C}^{(m)}) \| \notag \\ 
&
\textstyle
+\frac{1}{2\gammab{}}\sum\limits_{i=1}^N\max(0,\xi_i-\gammab{}(1-y_i\rho_i))^2 
+\frac{1}{2\gammab{}}\sum\limits_{i=1}^N\max(0,\zeta_i-\gammab{} y_i\rho_i)^2,
\label{eq:phigamma_two} 
\end{align}
and $\gamma\in\Realp$. 
The gradient and the Hessian of $\varphi_{\gamma}$ with respect to $\rho$ 
can be obtained in a similar way to \Eqsref{eq:varphi_grad} and \eqref{eq:varphi_hess}. 
Thus we use the Newton method for the minimization \eqref{eq:phigamma_two}. 
The overall algorithm is analogous to Table \ref{tab:Algorithm1} with update equations \eqref{eq:update_hinge_one}-\eqref{eq:phigamma_two}.

\subsubsection{Optimization for general loss functions with constraints}
Here we generalize the above argument to a broader class of loss functions.
We assume that 
the dual of the loss function can be written by using twice differentiable convex functions $\ell_0$ and $h_j$ as  
\begin{align}
\ell^\ast(y_i,\rho_i) =  \min_{\tilde{\rho}} \{ \ell_0^\ast(y_i,(\rho_i,\tilde{\rho}_i)) \mid  h_j(y_i,(\rho_i,\tilde{\rho}_i)) \leq 0~(1\leq j \leq B) \},
\label{eq:ellast_const_inf}.
\end{align}
where $\tilde{\rho}_i \in \Real^{B'}$ is an auxiliary variable. 
An example is $\epsilon$-sensitive loss for regression that is defined as 
$$\ell(y,f) = \begin{cases} |f-y|-\epsilon &(\text{if}~|f-y|\geq \epsilon) \\ 0 &(\text{otherwise})\end{cases}. $$
By simple calculation, the dual function of the $\epsilon$-sensitive loss is given as 
\begin{align}
\ell^*(y,\rho) = \begin{cases} y \rho + |\rho| \epsilon  & (\text{if}~|\rho| \leq 1) \\ \infty &(\text{otherwise}) \\    \end{cases},
\end{align}
This is not differentiable, but is written as 
\begin{align}
\ell^*(y,\rho) = \min_{\tilde{\rho}}\{ y \rho + (\rho + 2\tilde{\rho})  \epsilon  \mid \rho - 1 \leq 0,~-1 - \rho \leq 0,~- \tilde{\rho} \leq 0,~-\rho - \tilde{\rho} \leq 0 \}.
\end{align}
Thus in this example, $\ell_0^\ast(y,(\rho,\tilde{\rho})) =  y \rho + (\rho + 2\tilde{\rho})$, $h_1(y,(\rho,\tilde{\rho})) = \rho - 1$, $h_2(y,(\rho,\tilde{\rho})) = - 1 - \rho $,
$h_3(y,(\rho,\tilde{\rho})) = - \tilde{\rho}$, and $h_4(y,(\rho,\tilde{\rho})) = -\rho - \tilde{\rho}$ in the formulation of \Eqref{eq:ellast_const_inf}.

The update equation becomes as following:
\begin{align}
\textstyle &(\alpha^{(t+1)},b^{(t+1)},\xi^{(t+1)})  = \!\!\!\!\!\! \\
&
\argmin\limits_{
\begin{subarray}{l}
\alpha\in\Real^{(MN)},b\in\Real\\ \xi\in\Realp^{NB}
\end{subarray}
}
\Biggl\{H_C(\alpha,b,\xi)+ \frac{1}{2\gammam{(t)}} \left( \|\alpha-\alpha^{(t)}\|_{\bar{K}}^2  
+ (b-b^{(t)})^2 
+ \|\xi-\xi^{(t)}\|^2\right)\Biggr\},
\label{eq:constraintPrimal}
\end{align}
where
$H_C$ is expressed in terms of maximum over $\rho\in\Real^N$, $\tilde{\rho}\in \Real^{NB'}$, $u \in \Real^{MN}$ as follows:
\begin{align*}
&H_C(\alpha,b,\xi)\\
&= \max_{\rho\in\Real^N,\tilde{\rho}\in\Real^{NB'}, u \in \Real^{MN} }\Biggl\{
-\sum_{i=1}^N \ell_0^\ast(y_i, (-\rho_i,\tilde{\rho}_i))  -\sum_{m=1}^M\delta_C^{m}(u_m)
-\sum_{m=1}^M\langle \alpha_m, \rho-u_m\rangle_{K_m} -b\sum_{i=1}^N\rho_i 
\\
&
~~~~~~~~~~~~~~~~~~~~~~-\sum_{i=1}^N \sum_{j=1}^B \xi_{i,j} h_j(y_i,(-\rho_i,\tilde{\rho}_i))
\Biggr\}.
\end{align*}
Note that the minimum of $H_C(\alpha,b,\xi)$ with respect to $\xi$ is the primal objective function $L(\bar{K}\alpha + b\boldone) + \phi_C(\alpha)$
because by exchanging $\min$ and $\max$ we have 
\begin{align*}
&\min_{\xi \in \Realp^{NB}} H_C(\alpha,b,\xi)\\
&= \max_{\begin{subarray}{c} \rho\in\Real^N \\\tilde{\rho}\in\Real^{NB'}, u \in \Real^{MN}~~ \end{subarray}} \inf_{\xi \in \Realp^{NB}}  \Biggl\{
-\sum_{i=1}^N \ell_0^\ast(y_i, (-\rho_i,\tilde{\rho}_i))  -\sum_{m=1}^M\delta_C^{m}(u_m)
-\sum_{m=1}^M\langle \alpha_m, \rho-u_m\rangle_{K_m} -b\sum_{i=1}^N\rho_i 
\\
&
~~~~~~~~~~~~~~~~~~~~~~~~~~~~-\sum_{i=1}^N \sum_{j=1}^B \xi_{i,j} h_j(y_i,(-\rho_i,\tilde{\rho}_i)) \Biggr\} \\
&= \max_{\rho\in\Real^N \tilde{\rho}\in\Real^{NB'}} \Biggl\{
-\sum_{i=1}^N \ell_0^\ast(y_i, (-\rho_i,\tilde{\rho}_i))  + \rho^\top(\bar{K}\alpha + b\boldone) \mid h_j(y_i,(-\rho_i,\tilde{\rho}_i))\leq 0 ~(\forall i,j) \Biggr\} \\
&~~~~~~+ \max_{ u \in \Real^{MN}} \left\{ -\sum_{m=1}^M\delta_C^{m}(u_m)
+ \sum_{m=1}^M \langle \alpha_m, u_m\rangle_{K_m}  \right\}
\\
&=L(\bar{K} \alpha + b \boldone) + \phi_C(\alpha).
\end{align*}
However it is difficult to directly optimize the primal objective.
Therefore we add the penalty term as in \Eqref{eq:constraintPrimal} and solve its dual problem.
We exchange the order of minimization and maximization as in the last section (see also~\Secref{sec:dal-prox})
and remove $\alpha,b,\xi$, and $u_m$ by explicitly minimizing or maximizing over them. 
Finally we obtain the following update equations.
\begin{align}
\textstyle
 \alpha_m^{(t+1)}&
\textstyle
=\prox(\alpha_m^{(t)} + \gammam{(t)}\rho^{(t)} | \phi_{\gamma^{(t)}C}^{(m)}), 
\\
\textstyle
b^{(t+1)} 
&=b^{(t)}+\gammab{(t)}\sum_{i=1}^N\rho_i^{(t)}
\\
\textstyle
\xi_{i,j}^{(t+1)} 
&\textstyle
=\max(0,\xi_{i,j}^{(t)}-\gamma^{(t)}h_j(y_i,(-\rho_i^{(t)},\tilde{\rho}_i^{(t)}))).
\label{eq:update_general_loss_two}
\end{align}
where 
$(\rho^{(t)},\tilde{\rho}^{(t)}) \in \Real^N \times \Real^{NB'}$ is the minimizer of the
function
 $\varphi_{\gamma^{(t)}}(\rho,\tilde{\rho};\alpha^{(t)},b^{(t)},\xi^{(t)})$
 defined as follows:
\begin{align} 
\textstyle
\varphi_{\gamma}(\rho,\tilde{\rho};\alpha,b,\xi) = & \textstyle
 \sum\limits_{i=1}^N \ell_0^\ast(y_i, (-\rho_i,\tilde{\rho}_i)) 
 +\frac{1}{2\gammab{}}(b+\gammab{} \sum\limits_{i=1}^N\rho_i)^2 
+ \sum\limits_{m=1}^M\frac{1}{2\gammam{}} \|\prox(\alpha_m + \gamma \rho | \phi_{\gamma C}^{(m)})\|_{K_m}^2 \notag \\ 
&
\textstyle
+\frac{1}{2\gammam{}}\sum\limits_{i=1}^N \sum\limits_{j=1}^B \max(0,\xi_{i,j}-\gammam{} h_j(y_i,(-\rho_i,\tilde{\rho}_i)))^2,
\label{eq:phigamma_general_two} 
\end{align}
and $\gamma \in\Realp$.
The gradient and the Hessian of $\varphi_{\gamma}$ with respect to $(\rho,\tilde{\rho})$ 
can be obtained in a similar way to \Eqsref{eq:varphi_grad} and \eqref{eq:varphi_hess}. 
Thus we use the Newton method for the minimization \eqref{eq:phigamma_two}. 
The overall algorithm is summarized in Table \ref{tab:Algorithm_generalreg_generalloss}. 

\begin{table*}[t]
\caption{Algorithm of \propalgo for a general regularization term and a general loss function}
\label{tab:Algorithm_generalreg_generalloss}
\begin{center}
\framebox{
  \begin{minipage}[t]{0.45\textwidth}
    \begin{tabbing}
      XX\=\kill
      1.\>Choose a sequence $\gammam{(t)}\to \infty$ as $t\to \infty$.\\
      2.\>Minimize the augmented Lagrangian with respect to $\rho$ and $\tilde{\rho}$:    \\
      \phantom{2.} 
$(\rho^{(t)},\tilde{\rho}^{(t)}) \!= \! \mathop{\argmin}_{\rho,\tilde{\rho}}\Big( \! \sum_{i}\ell_0^{\ast}(y_i,(-\rho_i,\tilde{\rho}_i))\!+\!  \frac{1}{2\gammam{(t)}}  \sum_m \|\prox(\alpha_m^{(t)} + 
\gammam{(t)} \rho | \phi_{\gammam{(t)} C}^{(m)})\|_{K_m}^2  \!\!$ \\
~~~~~~~~~~~~~~~~~~$ +\frac{1}{2\gammam{(t)}}\sum\limits_{i=1}^N \sum\limits_{j=1}^B \max(0,\xi_{i,j}^{(t)}-\gammam{} h_j(y_i,(-\rho_i,\tilde{\rho}_i)))^2 
+ \frac{1}{2\gammab{(t)}} (b^{(t)}+\gammab{(t)}\sum_i \rho_i )^2\Big).$ \\
      3.\>Update $\alpha_m^{(t+1)} \leftarrow \prox( \alpha_m^{(t)}+\gammam{(t)}\rho^{(t)}| \phi_{\gammam{(t)}C}^{m}),~
b^{(t+1)}  \leftarrow b^{(t)} + \gammab{(t)} \sum_i \rho_i^{(t)},$ \\
~~~~~$\xi_{i,j}^{(t+1)} 
\leftarrow \max(0,\xi_{i,j}^{(t)}-\gamma^{(t)}h_j(y_i,(\rho_i^{(t)},\tilde{\rho}_i^{(t)}))).$
 \\
      4.\>Repeat 2.~and 3.~until the stopping criterion is satisfied. 
    \end{tabbing}
 \end{minipage}
}
\end{center}
\end{table*}

\subsection{Technical details of computations}
\subsubsection{Back-tracking in the Newton update} 
\label{sec:Technicaldetailsofcomputations}
We use back-tracking line search with {\it Armijo's rule} to find a step size of the Newton method.
During the back-tracking, 
the computational bottle-neck is the computation of the norm $\|\rho + c \Delta \rho + \frac{\alpha_m}{\gamma_m}\|_{K_m}$ $(m=1,\dots,M)$
where $\Delta \rho$ is the Newton update direction and $0 <c \leq 1$ is a step size. 
However, the above computation is necessary only for the active kernels $M_+$; 
for example, the active kernels $M_+$ is included in $\{m\mid \|\rho + \frac{\alpha_m}{\gamma_m} \|_{K_m} > C$ \text{or} $\|\rho + \Delta\rho + \frac{\alpha_m}{\gamma_m} \|_{K_m} > C \}$
for block 1-norm MKL and 
$\{m\mid \|\rho \|_{K_m} > (1-\lambda) C$ \text{or} $\|\rho + \Delta\rho \|_{K_m} > (1-\lambda) C \}$
for Elastic-net MKL
because of the convexity of $\|\cdot \|_{K_m}$.
This reduces the computation time considerably.  

\subsubsection{Parallelization in the computation of $\big\|\alpha_m^{(t)} + \frac{\rho^{(t)}}{\gamma^{(t)}_m} \big\|_{K_m}$ } 
The computation of $\big\|\alpha_m^{(t)} + \frac{\rho^{(t)}}{\gamma^{(t)}_m} \big\|_{K_m}$ for $m=1,\dots,M$  
is necessary at every outer iteration. 
Thus when the number of kernels is large, 
its cost cannot be ignored.
Fortunately, the computation with respect to the $m$-th kernel is independent of the others and the outer loop is easily parallelizable. 
In our experiments, we implemented the parallel computing by OpenMP 
in mex-files of our Matlab$^{\scriptsize \textcircled{\tiny{R}}}$ code.
We observed twenty or thirty percent improvement of overall computational time
by the parallelization.

\section{Generalized block-norm formulation}
\label{sec:general}
Motivated by the recent interest in non-sparse regularization for MKL,
we consider a generalized block norm formulation in this section. The
proposed formulation uses a concave function $g$ and it subsumes
previously proposed MKL models, such as $\ell_p$-norm
MKL~\citep{KloBreSonLasMueZie09} and Elastic-net
MKL~\citep{arXivSparsAcc:Tomioka+Suzuki:2009}, as different choices of the
concave function $g$. Moreover, using a convex
upper-bounding technique~\citep{PalWipKreRao06} we show that the proposed
formulation can be written also as a Tikhonov regularization
problem~\eqref{eq:mkl-tikhonov-l1}. Thus the solution can be easily
mapped to kernel weights as in previous approaches. We
extend SpicyMKL algorithm for the generalized
formulation in \Secref{sec:spicy-gen}. Furthermore we preset an efficient one-step optimization
procedure for Elastic-net MKL and $\ell_p$-norm MKL in \Secref{sec:onestepopt}.

\subsection{Proposed formulation}
We consider the following generalized block-norm formulation:
\begin{align}
\label{eq:mkl-block-gen}
  \minimize_{f_1\in\calH_1,\ldots,
 f_M\in\calH_M,b\in\Real}&\sum_{i=1}^N\ell\left(y_i,\textstyle\sum_{m=1}^Mf_m(x_i)+b\right)+C\sum_{m=1}^Mg(\|f_m\|_{\calH_m}^2),
\end{align}
where $g$ is a non-decreasing concave function defined on the
nonnegative reals, and we assume that $\tilde{g}(x)=g(x^2)$ is a convex
function of $x$. For example, taking $g(x)=\sqrt{x}$ gives the block
1-norm MKL in \Eqref{eq:mkl-block1-1} and taking $g(x)=x^{q/2}/q$ gives the
block $q$-norm MKL~\citep{KloBreSonLasMueZie09,NatDinRamBhaBenRam09}.
Moreover by choosing $g(x)=(1-\lambda)\sqrt{x}+\frac{\lambda}{2}x$, we
obtain the following Elastic-net
MKL~\citep{arXivSparsAcc:Tomioka+Suzuki:2009}:
\begin{align}
\label{eq:mkl-block-elastic}
  \minimize_{f_1\in\calH_1,\ldots,
 f_M\in\calH_M,b\in\Real}&\sum_{i=1}^N\ell\left(y_i,\textstyle\sum_{m=1}^Mf_m(x_i)+b\right)+C\sum_{m=1}^M\left((1-\lambda)\|f_m\|_{\calH_m}+\frac{\lambda}{2}\|f_m\|_{\calH_m}^2\right),
\end{align}
which reduces to the block 1-norm regularization \eqref{eq:mkl-block1-1}
for $\lambda=0$ and the uniform-weight combination ($d_m=1$ ($\forall m$)
in \Eqref{eq:fixed-kernel-learning}) for $\lambda=1$.

 Note that the generalized
regularization term in the above formulation \eqref{eq:mkl-block-gen} is
separable into each kernel component; thus it can be more efficiently
handled than the squared regularization used in previous
studies~\citep{KloBreSonLasMueZie09,KloRucBar10}.

The first question we need to answer is how the generalized block-norm
formulation~\eqref{eq:mkl-block-gen} is related to the Tikhonov
regularization formulation~\eqref{eq:mkl-tikhonov-l1}. This can be shown
using a convex upper-bounding technique; see~\citep{PalWipKreRao06}.  For any given concave function
$g$, the {\em concave conjugate} $g^\ast$ of $g$ is defined as follows:
\begin{align*}
 g^\ast(y)=\inf_{x\geq 0}\left( xy-g(x)\right).
\end{align*}
The definition of concave conjugate immediately implies that the
following inequality is true:
\begin{align*}
 g\left(\|f_m\|_{\calH_m}^2\right)&\leq \frac{\|f_m\|_{\calH_m}^2}{2d_m}-g^\ast\left(\frac{1}{2d_m}\right).
\end{align*}
Note that the equality is obtained by taking
$d_m=1/(2g'(\|f_m\|_{\calH_m}^2))$, where $g'$ is the derivative of $g$.
Therefore, the generalized block-norm
formulation~\eqref{eq:mkl-block-gen} is equivalent to the following Tikhonov
regularization problem:
\begin{align}
\label{eq:mkl-tikhonov}
  \minimize_{
\substack{
f_1\in\calH_1,\ldots,f_M\in\calH_M,\\
b\in\Real,\\
d_1\geq 0,\ldots,d_M\geq 0}}&\sum_{i=1}^N\ell\left(y_i,\textstyle\sum_{m=1}^Mf_m(x_i)+b\right)+\frac{C}{2}\sum_{m=1}^M\left(\frac{\|f_m\|_{\calH_m}^2}{d_m}+ h(d_m)\right),
\end{align}
where we define the regularizer $h(d_m)=-2g^\ast(1/(2d_m))$. It is easy
to obtain the regularizer $h$ corresponding to the block 1-norm MKL,
block $q$-norm MKL, and the Elastic-net MKL, and it is shown in
Table~\ref{tab:correspond}. 
See \citet{arXivRegSt:Tomioka+Suzuki:2011} for more detailed and generalized discussions about the relations between 
the regularizations on the RKHS norm $\{\|f_m\|_{\calH_m}\}_{m}$ and the kernel weight $\{d_m\}_{m}$.

Once we obtain the optimizer $\{f^*_m\}_m$ for the generalized block-norm formulation \eqref{eq:mkl-block-gen},
then we can easily recover the corresponding kernel weight by the relation
$$
d_m= \frac{1}{2g'(\|f_m^*\|_{\calH_m}^2)} = \frac{1}{2g'(\|\alpha_m^*\|_{K_m}^2)},
$$
where $\alpha_m^*$ is the optimal coefficient vector corresponding to $f^*_m$. 
For the Elastic-net MKL, for example, the kernel weight is recovered as 
$$
d_m= \begin{cases}0, & (\|\alpha_m^*\|_{K_m}=0), \\ \frac{\|\alpha_m^*\|_{K_m}}{ 1-\lambda + \lambda \|\alpha_m^*\|_{K_m}}, & (\text{otherwise}). \end{cases}
$$
Note that we have a freedom of a constant multiplication for the kernel weight.

\begin{table}[tb]
 \begin{center}
\caption{Correspondence of the concave function $g$ in
  \Eqref{eq:mkl-block-gen} and the regularizer $h$
  in \Eqref{eq:mkl-tikhonov}. For block $q$-norm MKL, the exponent $q$
  in the block norm formulation and the exponent $p$ in the Tikhonov
  regularization problem correspond as $p:=q/(q-2)$. $I_{[0,1]}$ denotes
  the indicator function of 
  the interval $[0,1]$; i.e., $I_{[0,1]}(x)=0$ (if $x\in[0,1]$), and
  $I_{[0,1]}(x)=\infty$ (otherwise). }
\label{tab:correspond}
  \begin{tabular}{c|c|c|c}
   MKL model         & $g(x)$       & $h(d_m)$   \\
\hline
\hline
   block 1-norm MKL  & $\sqrt{x}$  & $d_m$ \\
   block $q$-norm MKL & $\frac{1}{q}x^{q/2}$ & $\frac{1}{p}d_m^p$ \\
   Elastic-net MKL & $(1-\lambda)\sqrt{x}+\frac{\lambda}{2}x$ & 
$\frac{(1-\lambda)^2d_m}{1-\lambda d_m}$ \\
Uniform-weight MKL & $x/2$ & $I_{[0,1]}(d_m)$ 
  \end{tabular}
 \end{center}
\end{table}

\subsection{SpicyMKL for generalized block-norm formulation}
\label{sec:spicy-gen}
Now we are ready to extend SpicyMKL algorithm to the generalized
block-norm formulation~\eqref{eq:mkl-block-gen}. The original iteration
\eqref{eq:update_rho}-\eqref{eq:update_b} needs three modifications,
namely the proximity operator $\prox(\cdot|\phi_{C}^{(m)})$, the
conjugate regularizer $\delta_C^{(m)}$, and Moreau's envelope function $\Phi_C^{(m)}$.

First the proximity operator $\prox(\cdot|\phi_{C}^{(m)})$ is redefined
using the generalized regularization term $g(\|\cdot\|_{K_m}^2)$ as  follows:
\begin{align*}
 \prox(v_m|\phi_{C}^{(m)})&:=\argmin_{v_m'\in\Real^N}\left(
C g(\|v_m'\|_{K_m}^2)+\frac{1}{2}\|v_m'-v_m\|_{K_m}^2
\right).
\end{align*}
Note that the above minimization reduces to a one-dimensional
minimization along $v_m'=c v_m$. Let $\tilde{g}_C(x)=C g(x^2)$. In fact, due
to the Cauchy-Schwarz inequality, we have
\begin{align*}
 \tilde{g}_C(\|v_m'\|_{K_m})+\frac{1}{2}\|v_m'-v_m\|_{K_m}^2
&\geq \tilde{g}_C(\|v_m'\|_{K_m})+\frac{1}{2}(\|v_m'\|_{K_m}-\|v_m\|_{K_m})^2.
\end{align*}
The minimum is obtained when
\begin{align*}
 \|v_m'\|&=\prox(\|v_m\|_{K_m}| \tilde{g}_C)\\
 &=\argmin_{x\geq 0}\left(\tilde{g}_C(x)+\frac{1}{2}(x-\|v_m\|_{K_m})^2\right).
\end{align*}
For example, for Elastic-net MKL, the regularizer $\tilde{g}$ is written as
$\tilde{g}(x)=(1-\lambda)x+\lambda x^2/2$, and the one dimensional proximity operator
$\prox(\cdot|\tilde{g}_C)$ is obtained as follows:
\begin{align*}
\prox(x|\tilde{g}_C) &=
\begin{cases}
 0 & (\textrm{if $0\leq x\leq C(1-\lambda)$}),\\
\frac{x-C(1-\lambda)}{C\lambda+1} & (\textrm{otherwise}).
\end{cases}
\end{align*}
Therefore, the proximity operator $\prox(v_m|\phi_C^{(m)})$ is obtained
as follows:
\begin{align*}
 \prox(v_m|\phi_C^{(m)})&=
\begin{cases}
 0 & (\textrm{if $\|v_m\|_{K_m}\leq C(1-\lambda)$}),\\
\frac{\|v_m\|_{K_m}-C(1-\lambda)}{(C\lambda+1)\|v_m\|_{K_m}} v_m & (\textrm{otherwise}).
\end{cases}
\end{align*}

Second, the convex conjugate of the regularizer $\delta_C^{(m)}$ is
redefined as follows:
\begin{align}
 \delta_C^{(m)}(u_m)
 &:=\sup_{\alpha_m\in\Real^N}\left(\ip{u_m}{\alpha_m}_{K_m}-\tilde{g}_C(\|\alpha_m\|_{K_m})\right)\nonumber\\
&=\sup_{\alpha_m\in\Real^N}\left(\|u_m\|_{K_m}\|\alpha_m\|_{K_m}-\tilde{g}_C(\|\alpha_m\|_{K_m})\right)\nonumber\\
\label{eq:delta-gen}
&=\tilde{g}_C^{\ast}(\|u_m\|_{K_m}).
\end{align}
For example, for the same Elastic-net regularizer, the convex conjugate
$\tilde{g}_C^{\ast}(y)$ is obtained as follows:
\begin{align}
 \tilde{g}_C^{\ast}(y)&=\sup_{x\geq
 0}\left(xy-C(1-\lambda)x-\frac{C\lambda }{2}x^2\right)\nonumber\\
&=
\label{eq:elastic-conj}
\begin{cases}
0 & (\textrm{if $0\leq y\leq C(1-\lambda)$}),\\
\frac{(y-C(1-\lambda))^2}{2C\lambda} & (\textrm{otherwise}).
\end{cases}
\end{align}

Third the envelope function $\Phi_C^{(m)}$ is redefined in terms of the
above new conjugate regularizer $\delta_C^{(m)}$ as follows:
\begin{align*}
 \Phi_C^{(m)}(v_m)&:=\min_{v_m'\in\Real^N}\left(
\tilde{g}^{\ast}_C(\|v_m'\|_{K_m})+\frac{1}{2}\|v_m'-v_m\|_{K_m}^2
\right)\\
&=\min_{v_m'\in\Real^N}\left(
\tilde{g}^{\ast}_C(\|v_m'\|_{K_m})+\frac{1}{2}(\|v_m'\|_{K_m}-\|v_m\|_{K_m})^2
\right)\\
&=\widehat{\tilde{g}_C^\ast}(\|v_m\|_{K_m}),
\end{align*}
where  we again used Cauchy-Schwarz inequality in the
second line, and $\widehat{\tilde{g}_C^\ast}$ is the Moreau's envelope
function of $\tilde{g}^\ast_C$ as follows:
\begin{align*}
\widehat{\tilde{g}_C^\ast}(y) =\min_{y'\geq 0}\left(\tilde{g}_C^\ast(y')+\frac{1}{2}(y'-y)^2\right).
\end{align*}

With the above three modifications, the SpicyMKL iteration is rewritten
as follows:
\begin{align}
\label{eq:update_rho_gen}
 \rho^{(t)}&:=\argmin_{\rho\in\Real^N}\left(L^\ast(-\rho)+\frac{1}{\gamma^{(t)}}\sum_{m=1}^M\widehat{\tilde{g}^{\ast}_C}(\|\alpha_m+\gamma^{(t)}\rho\|_{K_m})+\frac{1}{2\gamma^{(t)}}(b+\gamma^{(t)}\textstyle\sum\limits_{i=1}^N\rho_i)^2\right),\\
\alpha_m^{(t+1)}&=\prox\left(\alpha_m^{(t)}+\gamma^{(t)}\rho^{(t)}\Big|\phi_{\gamma^{(t)}C}^{(m)}\right)\qquad(m=1,\ldots,M),\nonumber\\
b^{(t+1)}&=b^{(t)}+\gamma^{(t)}\sum_{i=1}^N\rho_i^{(t)}.\nonumber
\end{align}
Note that if $\tilde{g}_C^\ast(0)=0$ (this is the case if
$\tilde{g}_C(0)=0$), the envelope function
$\widehat{\tilde{g}_C^\ast}(\cdot)$ in \eqref{eq:update_rho_gen} becomes
zero whenever the norm $\|\alpha_m+\gamma^{(t)}\rho\|_{K_m}$ is
zero. Thus the inner objective \eqref{eq:update_rho_gen} inherits the
computational advantage provided by sparsity of the 1-norm SpicyMKL
presented in \Secref{sec:MinALFunc}. Moreover, the gradient and Hessian of the
generalized inner objective~\eqref{eq:update_rho_gen} can be computed in
a similar manner as \eqref{eq:varphi_grad} and \eqref{eq:varphi_hess}.

A generalization to the general loss functions with constraints is also straightforward;
we only need to replace $\|\prox(\alpha_m + \gamma \rho | \phi_{\gamma C}^{(m)})\|_{K_m}^2$ in \Eqref{eq:phigamma_general_two} 
(the definition of $\varphi_{\gamma}(\rho,\tilde{\rho};\alpha,b,\xi)$) 
with $\widehat{\tilde{g}^{\ast}_C}(\|\alpha_m+\gamma \rho\|_{K_m})$.

\paragraph{Implementation of general loss and regularization}
Our Matlab implementation is suited to general loss and block-norm regularization functions.
The code runs under general settings
if one plugs in scripts of the following information:
the primal and dual functions of the loss, the gradient and Hessian of the dual loss,
the primal and dual of the regularization function, the corresponding proximity operator and the derivative of the proximity operator.

\subsection{One step optimization for a regularization function with smooth
 dual} 
\label{sec:onestepopt}
When the convex conjugate $\delta_{C}^{(m)}$ of the regularization term
$\phi_{C}^{(m)}$ is smooth (twice differentiable), 
the optimization needs only {\it one} step iteration of the outer loop. 
We illustrate the optimization method for smooth dual function in a generalized formulation and 
show two useful examples, Elastic-net MKL for $\lambda>0$ and block
$q$-norm MKL for $q>1$. 

The primal problem \Eqref{eq:MKL_primal} is equivalent to the following dual problem 
by Fenchel's duality theorem \cite[Theorem 31.2]{Book:Rockafellar:ConvexAnalysis}:
\begin{align}
\textstyle 
\maximize \limits_{\begin{subarray}{c}
\rho\in\Real^{N}
\\ \boldone^\top \rho = 0
\end{subarray}}
\textstyle 
\bigg(
&-L^\ast(-\rho)-\sum\limits_{m=1}^M {\delta_{C}^{(m)}}(\rho)  \bigg), 
\label{eq:normaldualprob}
\end{align}
where $L^\ast$ and $\delta_{C}^{(m)}$ are the convex conjugate of $L$ and $\phi_{C}^{(m)}$.
Note that the constraint $\boldone^\top \rho = 0$ is due to the bias term $b$ so that 
if there is no bias term this constraint is removed. 
Using the expression for the convex conjugate $\delta_C^{(m)}$ in
\eqref{eq:delta-gen}, \Eqref{eq:normaldualprob} is rewritten as follows:
\begin{align}
\textstyle 
\maximize \limits_{\begin{subarray}{c}
\rho\in\Real^{N}
\\
\boldone^\top \rho = 0
\end{subarray}}
\textstyle 
\bigg(
&-L^\ast(-\rho)- \sum\limits_{m=1}^M \tilde{g}_C^{\ast}(\|\rho \|_{K_m}) \bigg). \nonumber
\end{align}
Thus if $L^\ast$ and $\tilde{g}^{\ast}$ are differentiable, 
the optimization problem can be easily solved by gradient descent or Newton method with the constraint $\boldone^\top \rho = 0$.
By a simple calculation, we obtain the gradient and Hessian of the dual regularization term as 
\begin{align}
&\nabla_{\rho} \tilde{g}_C^{\ast}\left(\|\rho \|_{K_m}\right) = \frac{K_m \rho}{\|\rho\|_{K_m}} \frac{\dd \tilde{g}_C^\ast(y)}{\dd y} \Bigg|_{y=\|\rho \|_{K_m}}, \label{eq:gradientgtil} \\
&\nabla \nabla^\top_{\rho} \tilde{g}_C^{\ast}\left(\|\rho \|_{K_m}\right) = \left(\frac{K_m}{\|\rho\|_{K_m}} - \frac{K_m \rho \rho^\top K_m}{\|\rho\|_{K_m}^3} \right) 
\frac{\dd \tilde{g}_C^\ast(y)}{\dd y} \Bigg|_{y=\|\rho \|_{K_m}}
+
\frac{K_m \rho  \rho^\top K_m}{ \|\rho\|_{K_m}^2} \frac{\dd^2 \tilde{g}_C^\ast(y)}{\dd y^2} \Bigg|_{y=\|\rho \|_{K_m}}.
\label{eq:Hessgtil}
\end{align}
To deal with the constraint $\boldone^\top \rho = 0$, we added a penalty function $C_b(\boldone^\top \rho)^2$ to the objective function 
in our numerical experiments with large $C_b$ (in our experiments we used $C_b=10^5$).

The primal optimal solution can be computed from the dual optimal solution as follows.
Given the dual optimal solution $\rho^\star$, the primal optimal solution $\alpha^\star = (\alpha_1^{\star\top},\dots,\alpha_M^{\star\top})^{\top}$ and $b^\star$
satisfy
$$\bar{K} \alpha^\star +  \boldone b^\star= \nabla_{\rho}L^\ast(\rho^\star),
$$
see \cite[Theorem 31.3]{Book:Rockafellar:ConvexAnalysis}.
By KKT-condition, there is a real number $c$ such that 
\begin{align}
 \nabla_{\rho}L^\ast(\rho^\star) &= -  \sum_{m=1}^M \nabla_{\rho}  \tilde{g}_C^{\ast}(\|\rho \|_{K_m})\Big|_{\rho=\rho^\star} + c\boldone \notag \\ 
&=-  \sum_{m=1}^M \frac{K_m \rho^\star}{\|\rho^\star \|_{K_m}} \frac{\dd \tilde{g}_C^\ast(y)}{\dd y} \Bigg|_{y=\|\rho^\star \|_{K_m}} + c\boldone. \notag 
\end{align}
Therefore the primal optimal solution is given as 
$$\alpha_m^\star = \frac{\rho^\star}{\|\rho^\star \|_{K_m}} \frac{\dd \tilde{g}_C^\ast(y)}{\dd y} \Bigg|_{y=\|\rho^\star \|_{K_m}},~~~b^\star = c.$$

When $L^\ast$ is not differentiable (e.g., hinge loss), 
combination of the techniques of this subsection and \Secref{sec:MinALfunctionGeneral}
can resolve the non-differentiability of $L^\ast$.

In the following we give two examples; Elastic-net and the block
$q$-norm regularization ($q>1$). 

\subsubsection{Efficient optimization of Elastic-net MKL}
\label{sec:onestepopt_elast}

In Elastic-net MKL, the regularization term is
$
\tilde{g}_C(x) = C(1-\lambda)x + \frac{C\lambda}{2}x^2
$.
The convex conjugate of $\tilde{g}_C$ is given in
\Eqref{eq:elastic-conj}.
Therefore
$$
\frac{\dd \tilde{g}_C^\ast(y)}{\dd y} = 
\begin{cases} 
0  & (|y| \leq C(1-\lambda)),\\
\frac{y-C(1-\lambda)}{C\lambda}  &(\text{otherwise}),
\end{cases}
~~~~~~~
\frac{\dd^2 \tilde{g}^\ast(y)}{\dd y^2} = 
\begin{cases} 
0  & (|y| \leq C(1-\lambda)),\\
\frac{1}{C\lambda}  &(\text{otherwise}).
\end{cases}
$$
Substituting the gradient and the Hessian to \Eqref{eq:gradientgtil} and \Eqref{eq:Hessgtil},
we obtain the Newton method for the dual of Elastic-net MKL.
It should be noted that the gradient and Hessian need to be computed only on active kernels as in Section \ref{sec:MinALFunc},
i.e. 
$
\nabla_{\rho} \tilde{g}^{\ast}(\|\rho \|_{K_m}) = 0,~
\nabla \nabla_{\rho}^\top \tilde{g}^{\ast}(\|\rho \|_{K_m}) =0
$
for all $m$ such that $\|\rho \|_{K_m} \leq C(1-\lambda)$.

\subsubsection{Efficient optimization of block $q$-norm MKL}
When $q>1$, block $q$-norm MKL can be solved in one outer step.
The regularization term of the block $q$-norm MKL is written as
$
\tilde{g}_C(x) = \frac{C}{q} x^{q}.
$
By a simple calculation, we obtain the convex conjugate $\tilde{g}_C^{\ast}$ as:
$$
\tilde{g}_C^\ast(y) =C^{1-r}\frac{1}{r}y^r,
$$
where $r=\frac{q}{q-1}$. Note that $r>1$ when $q >1$.
Then we have  
$$
\frac{\dd \tilde{g}^\ast(y)}{\dd y} = C^{1-r}y^{r-1},
~~~~~~~
\frac{\dd^2 \tilde{g}^\ast(y)}{\dd y^2} = C^{1-r}(r-1) y^{r-2}.
$$
Therefore we can apply the Newton method to the dual of $\ell_p$-norm MKL 
using the formulae \eqref{eq:gradientgtil} and \eqref{eq:Hessgtil}.

\section{Relations with the existing methods}
\label{sec:relationExistMeth}

We briefly illustrate the relations between our approach and the existing methods.
Basically the existing methods (such as 
SILP \citep{JMLR:Sonnenburg+etal:2006},
SimpleMKL \citep{JMLR:Rakotomamonjy+etal:2008},
LevelMKL \citep{NIPS:Xu+etal:2008}, and HessianMKL \citep{NIPSWS:Chapelle:2008})
rely on the equation \citep{JMLR:MicchelliPontil:2005}:
\begin{equation*}
\left(\sum_{m=1}^M \|f_m\|_{\calH_m}\right)^2
= \inf_{d_m \geq 0, \sum_m d_m = 1} \left\{ \sum_{m=1}^M \frac{\|f_m\|_{\calH_m}^2}{d_m} \right\}.
\end{equation*}
An advantage of this formulation is that we have a smooth upper bound $\sum_{m=1}^M \frac{\|f_m\|_{\calH_m}^2}{d_m}$ of the non-smooth $\ell_1$ regularization. 
Moreover \Eqref{eq:mkl-norm} says that minimizing this upper bound with respect to $\{f_m\}_{m=1}^M$ under the constraint $f=\sum_{m=1}^M f_m$ for a given function $f$, 
the upper bound becomes the RKHS norm of the function $f$ in the RKHS $\calH_{\bar{k}(d)}$ corresponding to the averaged kernel $\bar{k}(d) =\sum_{m=1}^M d_m k_m$, i.e., 
\begin{align*}
\|f\|_{\calH_{\bar{k}(d)}}^2 = \inf_{f=\sum_{m=1}^M f_m} \left\{ \sum_{m=1}^M \frac{\|f_m\|_{\calH_m}^2}{d_m} \right\},
\end{align*}
(see \citet{TAMS:Aronszajn:1950} for the proof).
Thus we don't need to consider $M$ functions $\{f_m\}_{m=1}^M$, instead we only need to deal with one function $f$ on the averaged kernel function $\bar{k}(d)$.
That is, the MKL learning scheme can be converted to 
\begin{align*}
&  \inf_{f_1\in\calH_1,\ldots,
 f_M\in\calH_M} \sum_{i=1}^N\ell\left(y_i,\textstyle\sum_{m=1}^Mf_m(x_i) \right)+C\left(\sum_{m=1}^M \|f_m\|_{\calH_m}\right)^2 \\
= &
  \inf_{d_m\geq 0,\sum_{m} d_m = 1} ~
\left\{  \inf_{f \in \calH_{\bar{k}(d)}}
\sum_{i=1}^N\ell\left(y_i, f(x_i) \right)+C \|f\|_{\calH_{\bar{k}(d)}}^2 \right\}.
\end{align*}
One notes that fixing $\{d_m\}_{m=1}^M$ the problem of the inner minimization is a standard single kernel learning.
Thus the inner minimization is easily solved by using publicly available efficient solvers for a single kernel learning. 
Based on these relations, the existing methods consist of the following two parts, the single kernel learning part and the updating part of $\{d_m\}_m$:
\begin{enumerate}
\item Minimize $L(\tilde{K} \alpha ) + C \alpha^\top \tilde{K} \alpha $ where $\tilde{K} = \sum_{m=1}^M d_m K_m$, 
\item Update $\{d_m\}_{m=1}^M$ so that the objective function decreases,
\end{enumerate}
where the update step differs depending on methods. 

On the other hand, our approach is totally different from the existing approaches. 
We don't utilize the kernel weight to obtain a smoothed approximation of the objective problem.
Our approach utilize the proximal minimization update \eqref{eq:prox_min} that can be interpreted as a gradient descent of 
the Moreau's envelope of the objective function.
In general, for a convex function $f$, the Moreau's envelope $\hat{f}$ defined below is differentiable: 
\begin{align}
&\hat{f}(x) := \min_y \left\{ f(y) + \frac{1}{2\gamma} \|y-x\|^2 \right\}, \notag \\
&\nabla \hat{f}(x) = \frac{1}{\gamma} (x-y^*(x)) \in \nabla f(y^*(x))~~~~~\text{where $y^*(x) = \mathop{\arg\min}_y \left\{ f(y) + \frac{1}{2\gamma} \|y-x\|^2 \right\}.$}
\label{eq:fhatDerivative}
\end{align}
One can check that $\min \hat{f} = \min f$ and $\arg \min \hat{f} = \arg \min f$ because $f(x^*) \leq f(y^*(x^*)) + \frac{1}{2\gamma} \|y^*(x^*)-x^*\|^2
= \hat{f}(x^*) \leq f(x^*) + \frac{1}{2\gamma} \|x^*-x^*\|^2 =  f(x^*)$ for all $x$ and $x^*\in \arg \min f$, and if $x \notin \arg \min f$, we have $f(x^*) < \hat{f}(x)$.
Thus the minimization of $\hat{f}$ leads to the minimization of $f$.
Using the relation \Eqref{eq:fhatDerivative}, 
the proximal minimization update \eqref{eq:prox_min} corresponds to 
$$
x^{(t+1)} \leftarrow y^*(x^{(t)}) = x^{(t)} - \gamma \nabla \hat{f}(x^{(t)}) \in x^{(t)} - \gamma \nabla f(x^{(t+1)}).
$$
Therefore the updating rule is a gradient descent of the Moreau's envelope of the objective function (and the increment is also the subgradient of $f$ at the {\it next solution} $x^{(t+1)}$).
Fortunately the minimization required to obtain the Moreau's envelope is obtained by a smooth optimization procedure as described in Section \ref{sec:MinALFunc}.
More general and detailed discussions about the use of the Moreau's envelope for machine learning settings can be found in \citet{JMLR:Tomioka+Suzuki+Sugiyama:2011}.


Regarding the optimization of block $q$-norm MKL, \citet{KloRucBar10} considered 
a unifying regularization including block $q$-norm MKL where the regularization term is 
$$
C_1 \left(\sum_{m=1}^M \|f_m\|_{\calH_m}^q \right)^{\frac{2}{q}} + C_2 \sum_{m=1}^M \|f_m\|_{\calH_m}^2,
$$
(there is additional operation $(\cdot)^{\frac{2}{q}}$ at the $q$-norm term compared with our block $q$-norm MKL formulation, 
however there is one-to-one correspondence between both formulations as in the same reason described in the end of Section \ref{sec:learningkernelweight}).
They also noticed that the dual of the above unifying regularization is smooth unless $C_2 = 0$ and $q=1$,
and suggest to utilize L-BFGS to solve the dual problem. 
This corresponds to the special case of our setting $g(x) = (1-\lambda) x^{q/2} + \lambda x$ with $0\leq \lambda \leq 1$.
As shown in Section \ref{sec:onestepopt}, we can solve the MKL problem with this regularization by one step outer iteration because of the smoothness of its dual.

\section{Numerical experiments}
\label{sed:NumericalExp}
In this section, we experimentally confirm the efficiency of 
the proposed SpicyMKL 
on several  binary classification tasks from UCI and IDA machine learning repository 
\footnote{All the experiments were executed on Intel Xeon 3.33GHz with 48GB RAM.}. 
We compared our algorithm SpicyMKL  
to four state-of-the-art algorithms namely SILP \citep{JMLR:Sonnenburg+etal:2006}, SimpleMKL \citep{JMLR:Rakotomamonjy+etal:2008},
LevelMKL~\citep{NIPS:Xu+etal:2008} and HessianMKL \citep{NIPSWS:Chapelle:2008}. 
As for SpicyMKL, we report the results of hinge and logistic losses with block 1-norm regularization.
We also report the result of the one step optimization method for elastic-net regularization 
(the method described in \Secref{sec:onestepopt_elast}) with hinge and logistic losses. 
To distinguish the iterative method and the one step method, 
we call the one step optimization method for elastic-net regularization as ElastMKL.

\subsection{Performances on UCI benchmark datasets}
We used 5 datasets from the UCI repository~\citep{Asuncion+Newman:2007}: `Liver', `Pima', `Ionospher', `Wpbc', `Sonar'. 
The candidate kernels were
Gaussian kernels with 24 
different bandwidths (0.1 0.25 0.5 0.75 1 2 3 4 $\cdots$ 19 20) 
and polynomial kernels of degree 1 to 3.    
All of 27 
different kernel functions (24 Gaussian kernels and 3 polynomial kernels) were applied to individual variables as well as jointly over all the variables; 
i.e., in total we have $27\times(n+1)$  
candidate kernels, where $n$ is the number of variables.
All kernel matrices were normalized to unit trace ($K_m \leftarrow K_m/\mathrm{trace(}K_m)$), and were precomputed prior to running the algorithms. 
These experimental settings were borrowed from the paper \citep{JMLR:Rakotomamonjy+etal:2008} of SimpleMKL, 
but we used a larger number of kernels. 

For each dataset, we randomly chose 80\% of samples for training  
and the remaining 20\% for testing.  
This procedure was repeated 10 times. 
Experiments were run on 3 different regularization parameters $C = 0.005$, $0.05$ and $0.5$;
for SimpleMKL, LevelMKL and HessianMKL, the regularization parameter was converted by \Eqref{CCdashCorrespondence}.
We employed 
the {\it relative duality gap}, $(\text{primal obj} - \text{dual obj})/\text{primal obj}$ with tolerance 0.01
as the stopping criterion for all algorithms. 
The primal objective for SpicyMKL and ElastMKL can be computed by 
using $\alpha^{(t)}$ and $b^{(t)}$.   
In order to compute the dual objective of SpicyMKL and ElastMKL, 
we project the multiplier vector $\rho$ to the equality constraint
$\tilde{\rho}=\rho - \boldone(\sum{\rho_i})/N$
and in addition, for SpicyMKL with block 1-norm regularization, 
we project $\tilde{\rho}$ to the domain of $\delta_C^{(m)}$ (\Eqref{eq:Aphistar}) 
by $\tilde{\rho}' = \tilde{\rho}/\max\{\max_m\{\|\tilde{\rho}\|_{K_m}/C\},1\}$.
Then we compute the dual objective function as $-L^{\ast}(-\tilde{\rho})$. 
The same technique 
can be found in \citet{ISPL:Tomioka+Sugiyama:2009} and \citet{WriNowFig09}.
The primal objective of SILP, SimpleMKL, HessianMKL and LevelMKL is computed as $\sum_i \alpha_i - \frac{1}{2} \sum_{i,j} \alpha_i \alpha_j \sum_m d_m K_m(x_i,x_j)$
and the dual objective as $\sum_i \alpha_i - \frac{1}{2} \max_m \sum_{i,j} \alpha_i \alpha_j K_m(x_i,x_j)$ where 
$\alpha_i$ is the dual variable of SVM solved at each iteration and $d_m$ is the kernel weight (see \citet{JMLR:Sonnenburg+etal:2006} and \citet{JMLR:Rakotomamonjy+etal:2008}).
 

For SpicyMKL and ElastMKL, we report the result from two loss functions; 
the hinge loss and the logistic loss. 
We used $\lambda=0.5$ for ElastMKL. 
For SILP, SimpleMKL, LevelMKL and HessianMKL, we used the hinge loss. 
For SILP, we used Shogun implementation\footnote{http://www.shogun-toolbox.org} written in C++, and for SimpleMKL\footnote{http://asi.insa-rouen.fr/enseignants/\char"7E arakotom/code/mklindex.html}, 
LevelMKL\footnote{http://appsrv.cse.cuhk.edu.hk/\char"7E zlxu/toolbox/level\_mkl.html} 
and HessianMKL\footnote{http://olivier.chapelle.cc/ams/}, we used publicly available Matlab$^{\scriptsize \textcircled{\tiny{R}}}$ codes. 
We replaced the Mosek$^{\scriptsize \textcircled{\tiny{R}}}$ solver with the CPLEX$^{\scriptsize \textcircled{\tiny{R}}}$ solver for the linear programming and the quadratic programming 
required inside LevelMKL.

\begin{figure*}[t]
\begin{center}
\hspace{-1cm}
\includegraphics[width=16.5cm,clip]{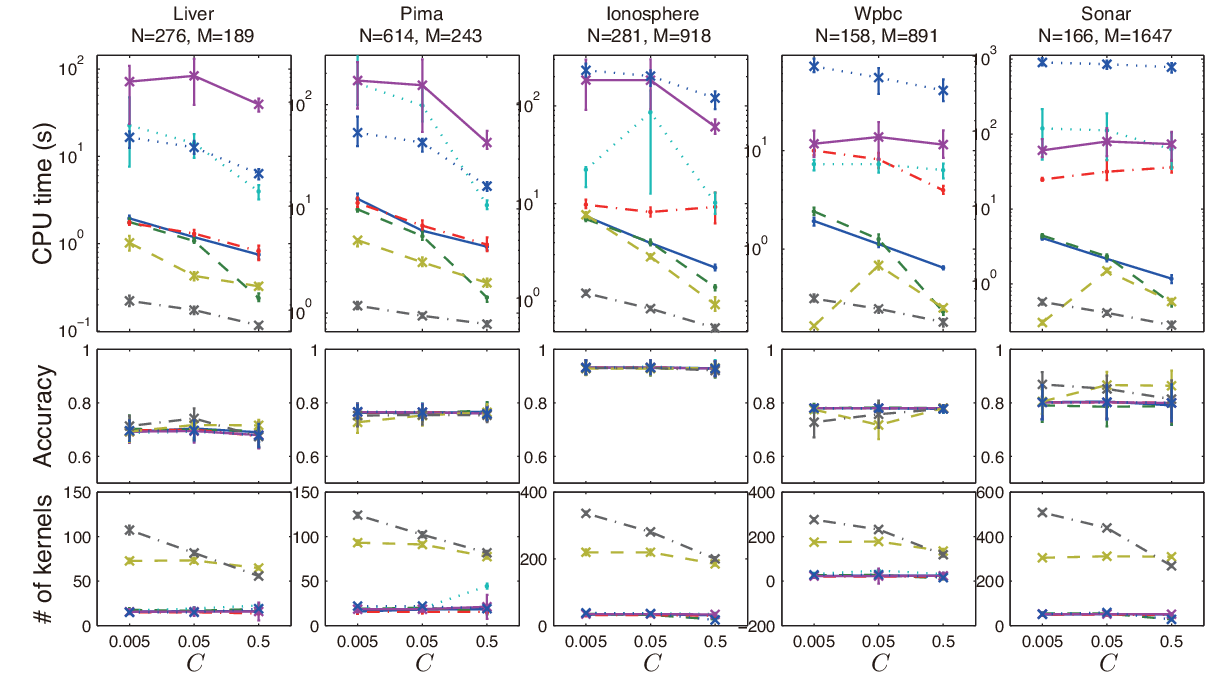} 
\includegraphics[width=13cm,clip]{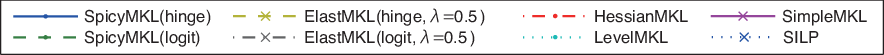} 
\vspace{-2.5mm}
\caption{Mean and standard deviation of performance measures for each MKL method for UCI datasets.} 
\label{fig:UCIdata}
\end{center}
\end{figure*}

The performance of each method is summarized in Figure~\ref{fig:UCIdata}. 
The average CPU time, 
test accuracy,
and 
final number of active kernels, are shown from top to bottom.  
In addition, standard deviations are also shown.
We can see that SpicyMKL tends to be faster than 
SILP (by factors of 4 to 670) and SimpleMKL (by factors of 6 to 60) and LevelMKL (by factors of 3 to 50),  
and faster than HessianMKL when the number of kernels $M$ is large (Ionospher, Wpbc \& Sonar). 
In all datasets, SpicyMKL becomes faster as the regularization parameter $C$ increases. 
This is because the larger $C$ becomes, the smaller 
the number of active kernels 
during the optimization. 
ElastMKL is even faster than SpicyMKL in all datasets. 
In particular, ElastMKL with the logistic loss shows the best performance among all methods.
This is because the one step optimization method explained in \Secref{sec:onestepopt} does not require iterative procedure for logistic loss.   
Accuracies of all methods for block 1-norm regularization are nearly identical. 
Thus from the classification accuracy, the block 1-norm MKL using logistic and hinge
loss seem to perform similarly.
The accuracy for elastic-net regularization varies slightly depending on problems.

SpicyMKL using the logistic loss tends to be faster than that using the hinge loss. 
This could be explained by
the strong convexity of the conjugate of the logistic loss, which is not
the case for the hinge loss.
In fact, when $L^{\ast}$ is strongly convex, the inner Newton method 
(minimization of $\varphi_{\gamma}$ with respect to $\rho$) converges rapidly.
Although the logistic loss is often faster to train, yet the accuracy is nearly identical to that of the hinge loss.

When the regularization is elastic-net, the hinge loss gives sparser solution than the logistic loss. 
The number of kernels selected under elastic-net regularization is much lager than that under block 1-norm regularization as expected,
and decreases as the strength of regularization $C$ increases.
Under block 1-norm regularization, the number of kernels slightly decreases as $C$ increases. 


In Table \ref{tab:averageIteration}, a comparison of the average numbers of outer iterations and SVM evaluations on the UCI datasets is reported. 
The existing methods (HessianMKL, LevelMKL, SimpleMKL, SILP) iterate the update of the kernel weight $\{d_m\}_{m=1}^M$ and 
the SVM evaluation of the objective function at the current weight as illustrated in Section \ref{sec:relationExistMeth}.
We refer to the combination of these two steps as one outer iteration.
Since SimpleMKL requires several SVM evaluations during the line search in gradient descent,
the number of outer iterations and that of SVM evaluations are different.
As for the other methods, both quantities are same. 
Obviously ElastMKL requires only one outer iteration, thus the result for ElastMKL is not reported in the table.
We can see that in all methods the number of outer iterations decreases as the regularization parameter $C$ increases. 
This is because as the regularization becomes strong, 
the required number of kernels becomes small and accordingly the intrinsic problem size becomes small.
SpicyMKL and HessianMKL require much smaller number of iterations than others.
This shows that these methods have fast convergence rate and thus the solution drastically approaches to the optimal one at each iteration.
Comparing the CPU time required by SpicyMKL and HessianMKL, SpicyMKL tends to require light computation per iteration especially for large number of kernels 
while HessianMKL requires QP to derive the update direction that is heavy for large number of kernels.

\begin{table}
\center
\caption{Average number of outer iterations. Average number of SVM evaluations is also shown in the brace for SimpleMKL.}
\label{tab:averageIteration}
\begin{tabular}{|c|c|rrrrrr|}
\hline 
Dataset& $C$  & \multicolumn{1}{c}{Spicy} & \multicolumn{1}{c}{Spicy} & \multicolumn{1}{c}{Simple} & \multicolumn{1}{c}{Hessian} & \multicolumn{1}{c}{Level} & \multicolumn{1}{c|}{SILP} \\ 
 & & \multicolumn{1}{c}{(hinge)} & \multicolumn{1}{c}{(logit)} &  &  &  &  \\ 
 \hline \hline 
    & 0.005&   22.6 &   20.1 &  200.3(3770.2) &   13.2 &  171.6 &  718.8  \\ 
Liver&0.05&   14.1 &   14.5 &  222.2(4800.0) &   11.8 &  138.0 &  517.2  \\ 
    & 0.5&    6.7 &    3.3 &  134.2(2200.5) &    9.4 &   50.1 &  237.8  \\ 
\hline 
    & 0.005&   31.5 &   21.5 &   76.2(1488.9) &   15.4 &  290.7 & 1020.1  \\ 
Pima&0.05&   13.6 &   13.9 &   74.9(1511.7) &   12.2 &  197.9 &  727.1  \\ 
    & 0.5&    7.9 &    3.2 &   24.6( 470.9) &   10.2 &   30.9 &  258.0  \\ 
\hline 
    & 0.005&   38.0 &   24.6 &  184.5(2864.1) &   11.3 &  100.9 & 1810.3  \\ 
Ionosphere&0.05&   18.3 &   17.1 &  188.6(2983.1) &   10.3 &  135.6 & 1580.1  \\ 
    & 0.5&    7.8 &    5.4 &   57.8(1374.4) &   14.4 &   65.5 &  972.0  \\ 
\hline 
    & 0.005&   24.0 &   24.4 &   45.4(1014.2) &   11.6 &   61.3 &  683.6  \\ 
Wpbc&0.05&   13.2 &   14.4 &   49.1(1134.7) &   10.7 &   60.8 &  514.2  \\ 
    & 0.5&    6.0 &    2.0 &   40.9( 964.4) &    7.2 &   55.4 &  367.0  \\ 
\hline 
    & 0.005&   35.2 &   27.2 &  158.3(2736.8) &   10.3 &  193.0 & 4484.6  \\ 
Sonar&0.05&   16.6 &   16.8 &  204.8(3535.3) &   10.6 &  194.2 & 4356.4  \\ 
    & 0.5&    7.7 &    4.2 &  173.3(3511.4) &   19.6 &  154.4 & 3862.3  \\ 
\hline 
\end{tabular}
\end{table}

\begin{figure*}[th]
\begin{center}
\subfigure[Evolution of Duality Gap]{
\includegraphics[height=4.7cm,clip]{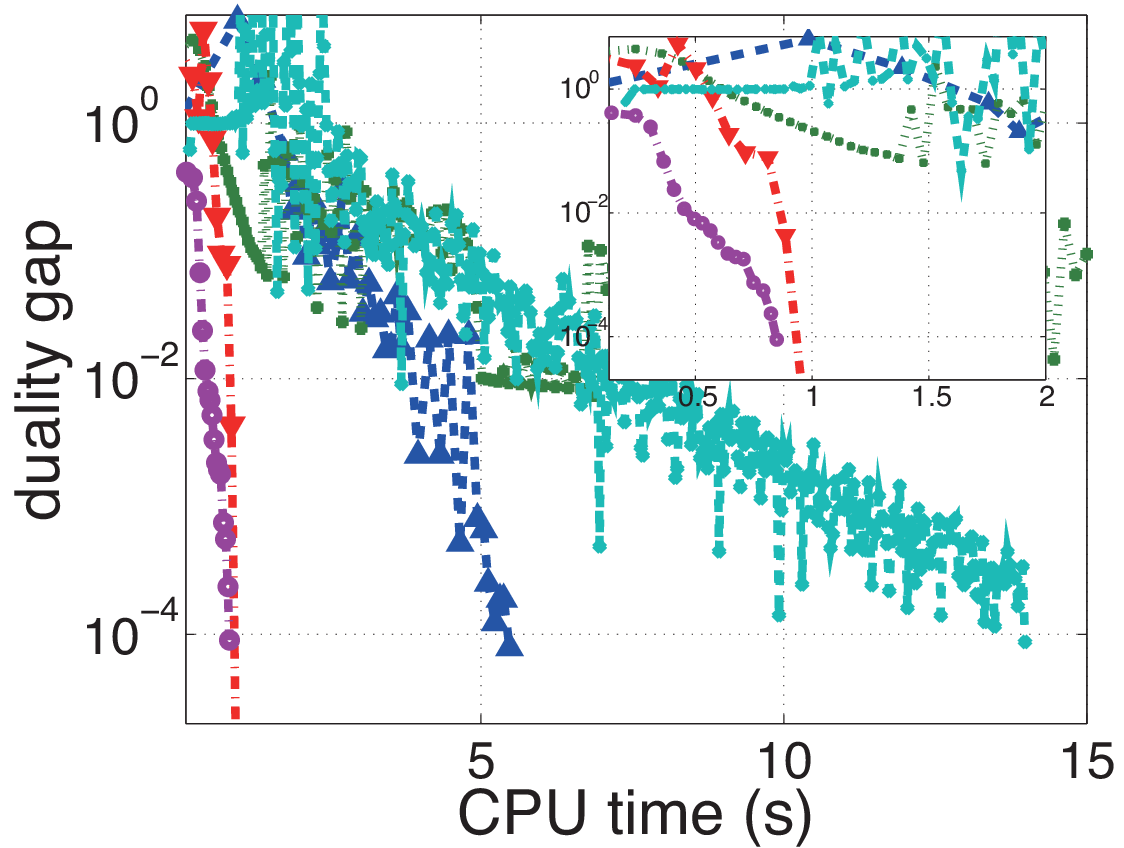}
\label{fig:TimeEvolutionDG}
}
\hspace{-2mm}
\subfigure[Evolution of \# of Active Kernels]{
\includegraphics[height=4.7cm,clip]{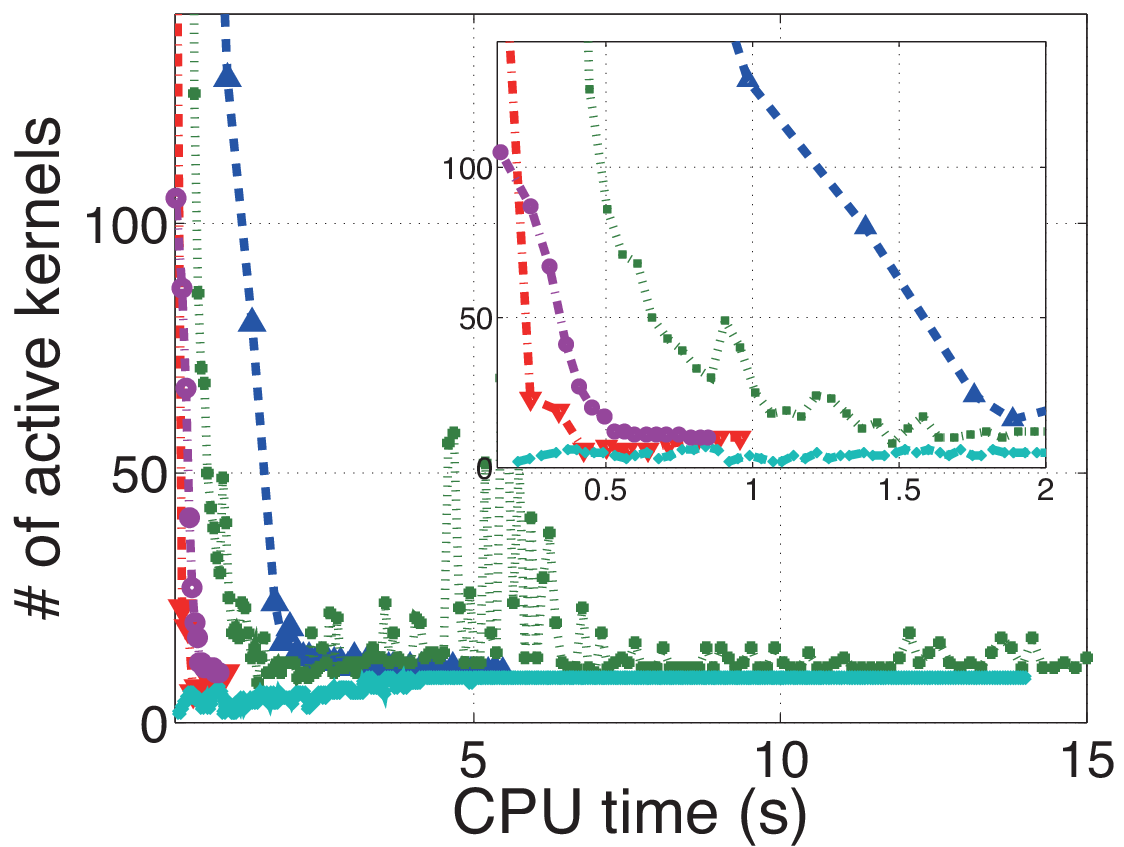}
\label{fig:TimeEvolutionAK}
} 
\hspace{-4mm}
\subfigure{
\mbox{\raisebox{7mm}{
\includegraphics[height=1.7cm,clip]{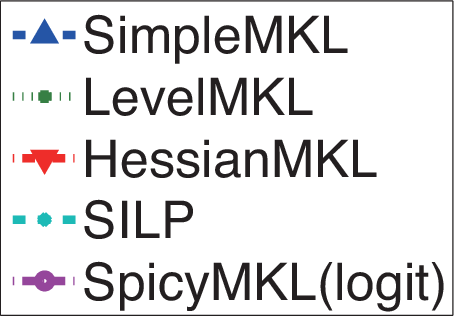} 
}}
} 
\caption{Duality gap and \# of active kernels against CPU time} 
\label{fig:TimeEvolution}
\end{center}
\end{figure*}

In Figure~\ref{fig:TimeEvolutionDG}, 
we plot the relative duality gaps against CPU time for a single run 
of SpicyMKL (with logistic loss and block 1-norm regularization), HessianMKL, LevelMKL and SimpleMKL
on the `Wpbc' dataset.
We can see that the duality gap of SpicyMKL drops rapidly and it is faster than linear. 
That supports the super-linear convergence of our method. 
HessianMKL decreases the duality gap quickly because HessianMKL is a second order method. 
The duality gap of LevelMKL gradually drops in the early stage, but
after several steps, the behavior of duality gap becomes unstable.
After all, the duality gap of LevelMKL did not drop below $10^{-2}$ in 500 iteration steps.
This is because the size of QP required in LevelMKL increases as the algorithm proceeds so that it gets hard to obtain a precise solution.
SILP shows fluctuations of the duality gap so that the duality gap does not decrease monotonically.
This is because the sequence of the solutions generated by cutting plane method shows oscillation behavior.
Figure~\ref{fig:TimeEvolutionAK} shows 
the number of active kernels as a function of the CPU time spent by the algorithm.  
Here we again observe rapid decrease in the number of kernels for SpicyMKL.
This reduces huge amount of computation per iteration. 
We see a relation between the speed of convergence and the number of kernels in SpicyMKL;
as the number of kernels decreases, the time spent per iteration becomes small. 



\subsection{Scaling against the sample size and the number of kernels}
\label{sec:ScalingSampleSize}
Here we investigate the dependency of CPU time on the number of kernels and the sample size. 
We used 4 datasets from IDA benchmark repository~\citep{MLJ:Ratch:2001}: 
`Ringnorm', `Splice', `Twonorm' and `Waveform'.
The same relative duality gap criterion with tolerance 0.01 was used. 
We generated a set of basis kernels by randomly selecting subsets of features and applying a Gaussian kernels with random width
$\sigma=5\chi^2+0.1$, where $\chi^2$ is a chi-squared random variable.
Here we report ElastMKL with $\lambda=0.1,0.5$ with hinge and logistic loss 
in addition to SpicyMKL with block 1-norm regularization with hinge and logistic loss, SimpleMKL, HessianMKL, LevelMKL and SILP. 

\begin{figure*}[t]
\begin{center}
\subfigure[Ringnorm]{
\label{fig:NumKernelScalingRing}
\includegraphics[height=4.7cm,clip]{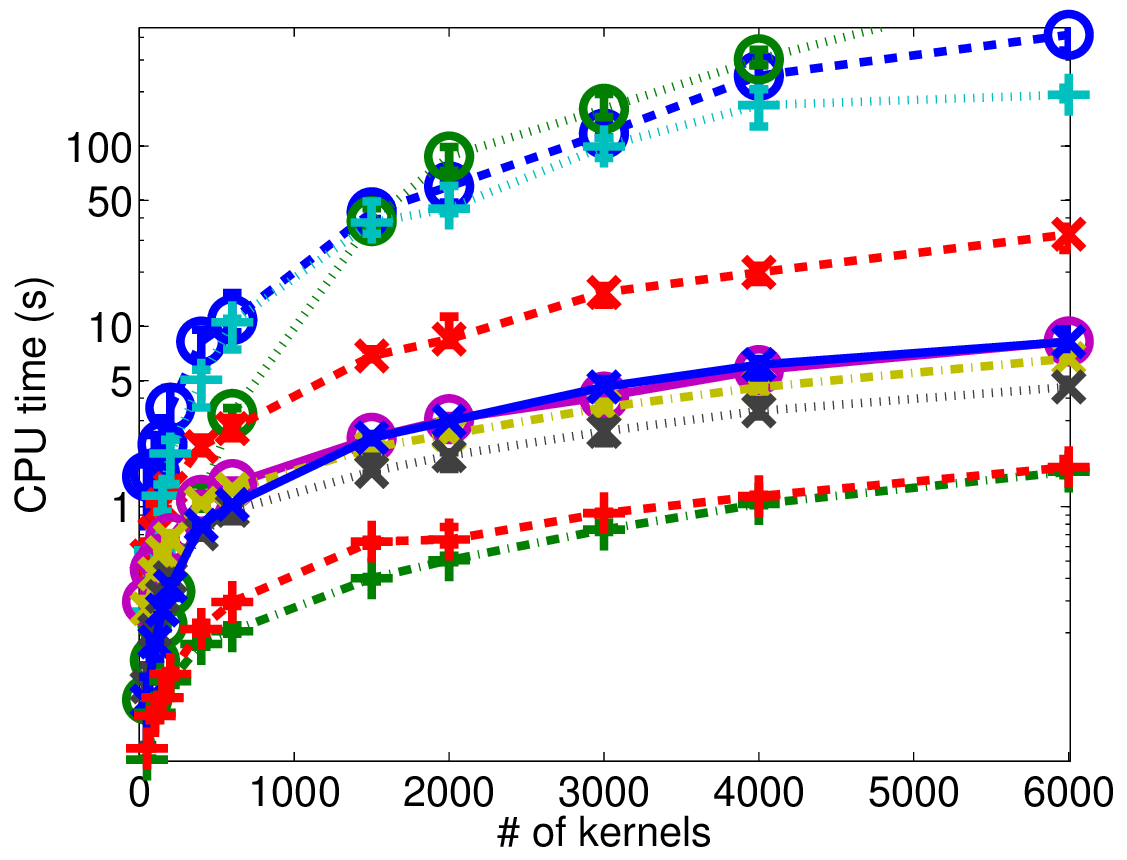}
}
\subfigure[Splice]{
\label{fig:NumKernelScalingSpli}
\includegraphics[height=4.7cm,clip]{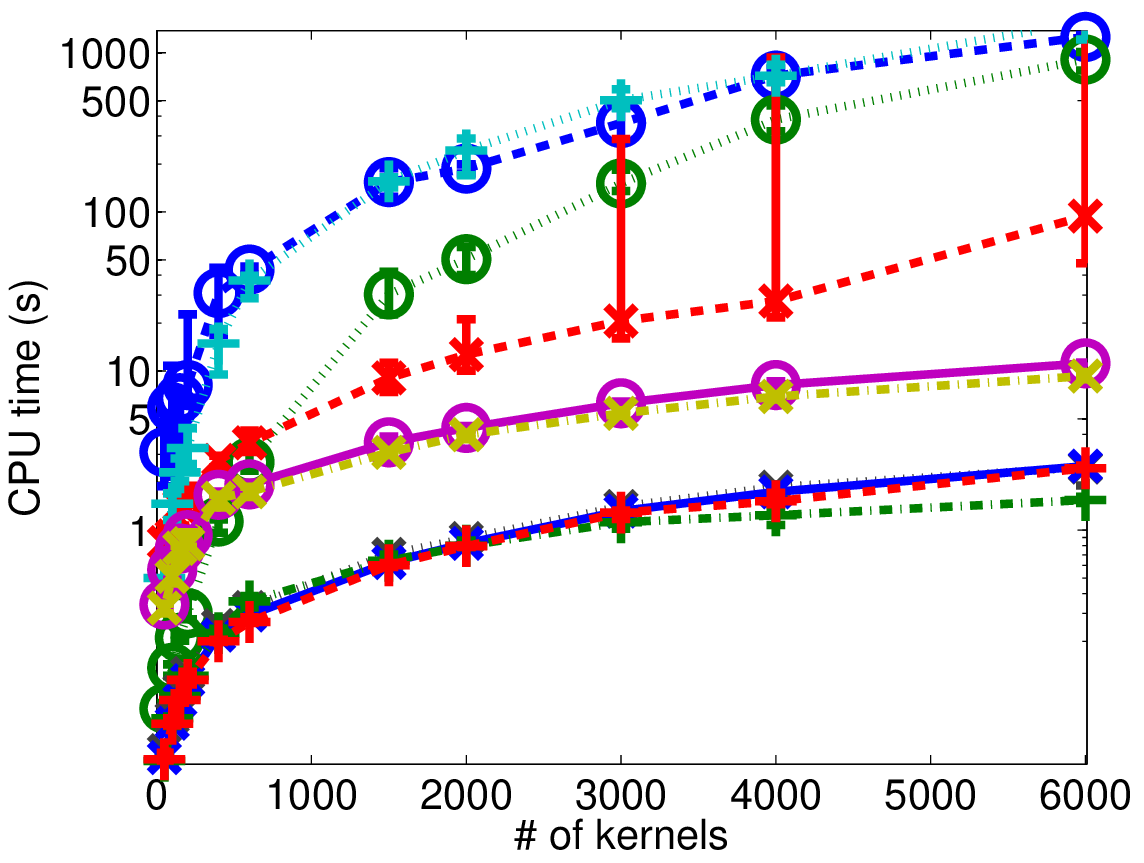}
}
\\
\vspace{0.5cm}
\subfigure[Twonorm]{ 
\label{fig:NumKernelScalingTwonorm}
\includegraphics[height=4.7cm,clip]{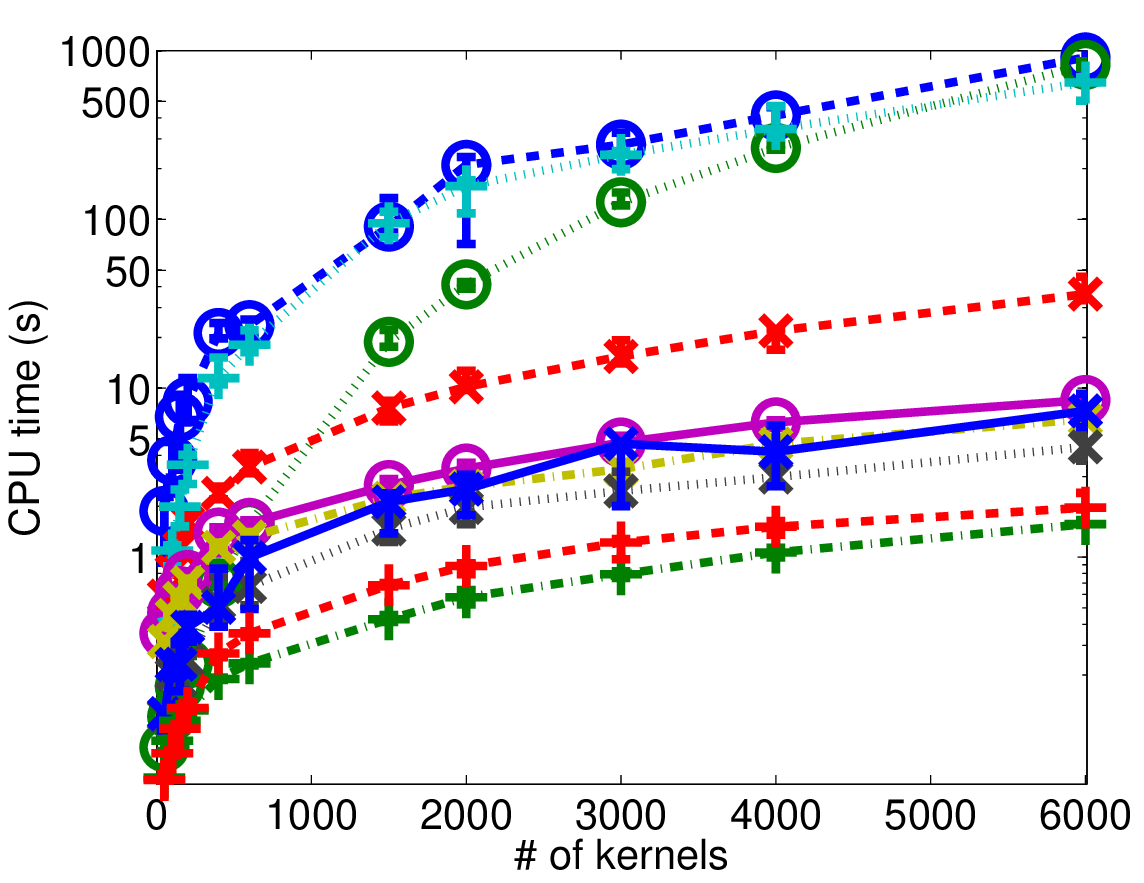}
}
\subfigure[Waveform]{ 
\label{fig:NumKernelScalingWaveform}
\includegraphics[height=4.7cm,clip]{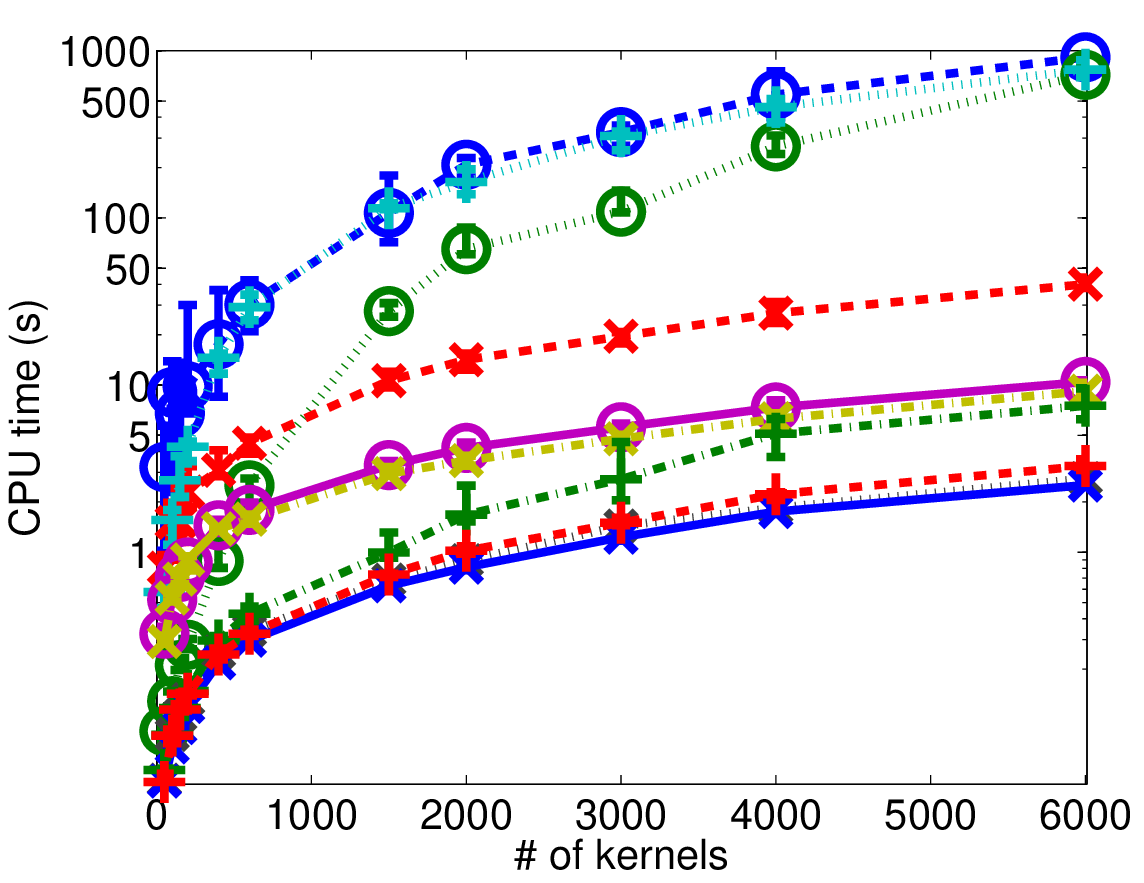}
}
\\
\vspace{0.2cm}
\subfigure{ 
\includegraphics[height=1.5cm,clip]{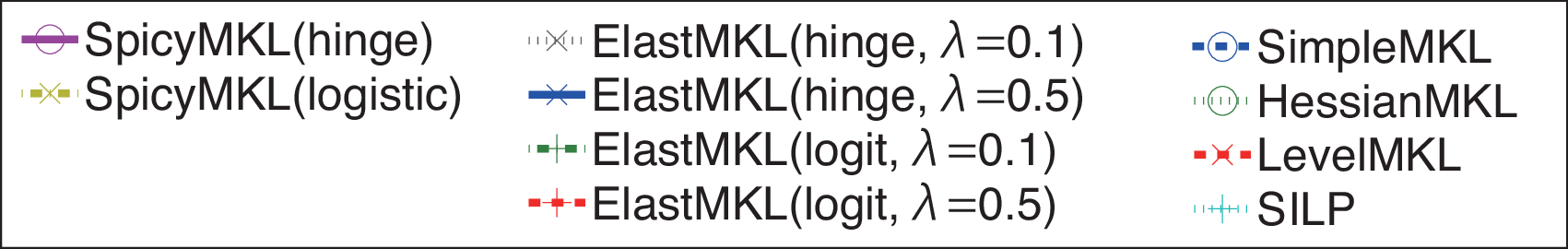}
}
\caption{CPU time as a function of the number of kernels}
\label{fig:NumKernelScaling}
\end{center}
\end{figure*}

Figure~\ref{fig:NumKernelScaling} the number of kernels is increased from 50 to 6000.
The graph shows the median of CPU times 
with 25 and 75 percentiles over 10 random train-test splitting 
where the size of training set was fixed to 200. 
We observe that 
for small number of kernels the CPU time of HessianMKL is the fastest among all methods with block 1-norm regularization. 
However when the number of kernels is greater than 1000, our method SpicyMKL is clearly the fastest among the methods with block 1-norm regularization. 
In fact, 
SpicyMKL is roughly 100 times faster 
than SimpleMKL, HessianMKL and SILP when the number of kernels is 6000.
LevelMKL scales almost same as SpicyMKL, but shows unstable performances in Splice dataset.
One reason is that LevelMKL shows the oscillation behavior found in Figure \ref{fig:TimeEvolutionDG}.
In particular, the quadratic programming required in LevelMKL often does not converge until the maximum number of iterations available in CPLEX.
The scaling property of SILP against the number of kernels is similar to that of SimpleMKL. 
Those methods do not have as good scaling property as SpicyMKL against the number of kernels.
Here again we observe that ElastMKL regularization is even faster than that with block 1-norm regularization,
and shows the best performance among all methods.

\begin{figure*}[t]
\begin{center}
\subfigure[Ringnorm]{
\label{fig:SampleScalingRing}
\includegraphics[height=4.7cm,clip]{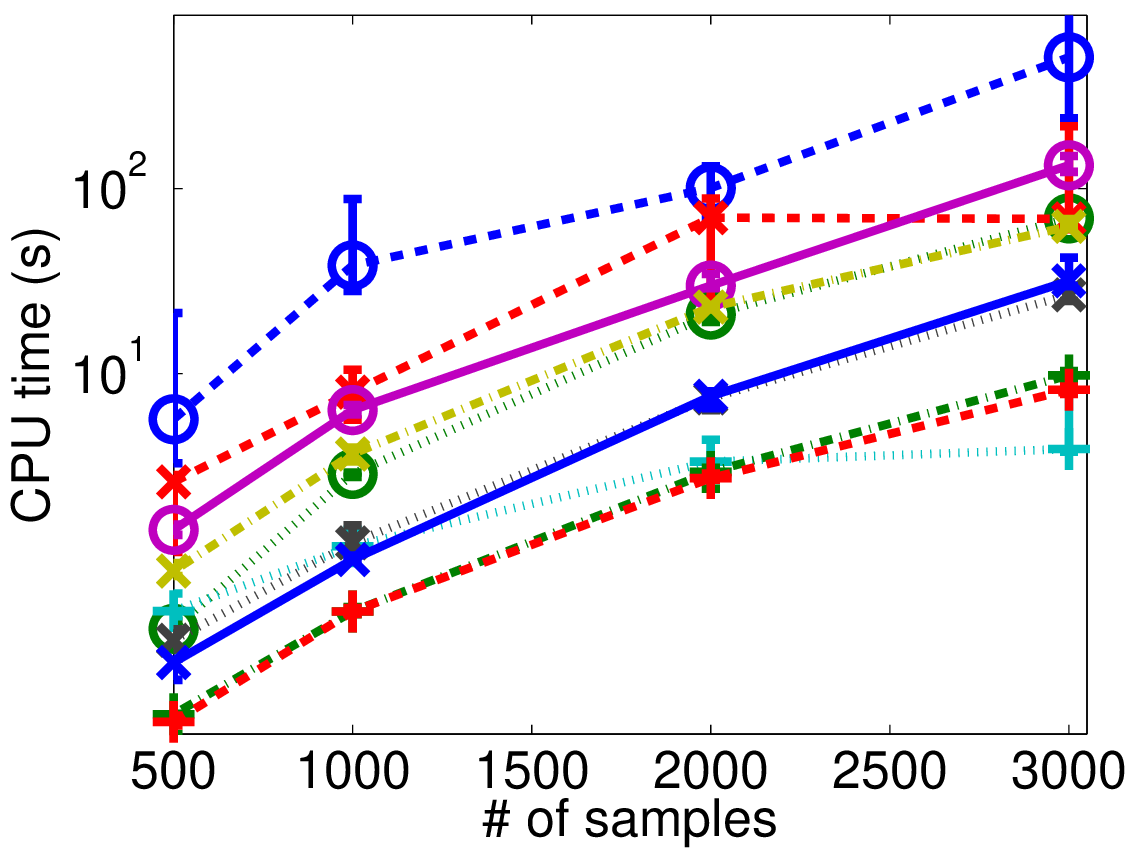}
}
\subfigure[Splice]{
\label{fig:SampleScalingSpl}
\includegraphics[height=4.7cm,clip]{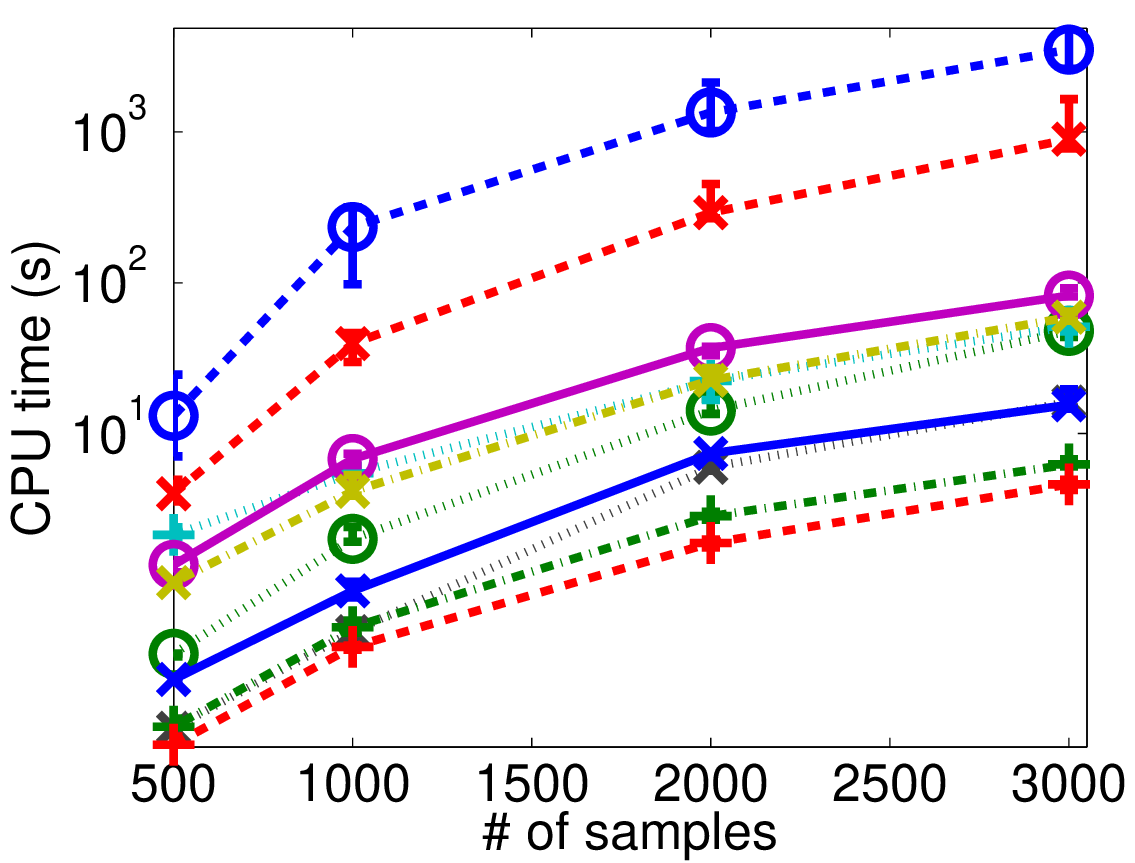}
}
\\
\vspace{0.5cm}
\subfigure[Twonorm]{
\label{fig:SampleScalingTwonorm}
\includegraphics[height=4.7cm,clip]{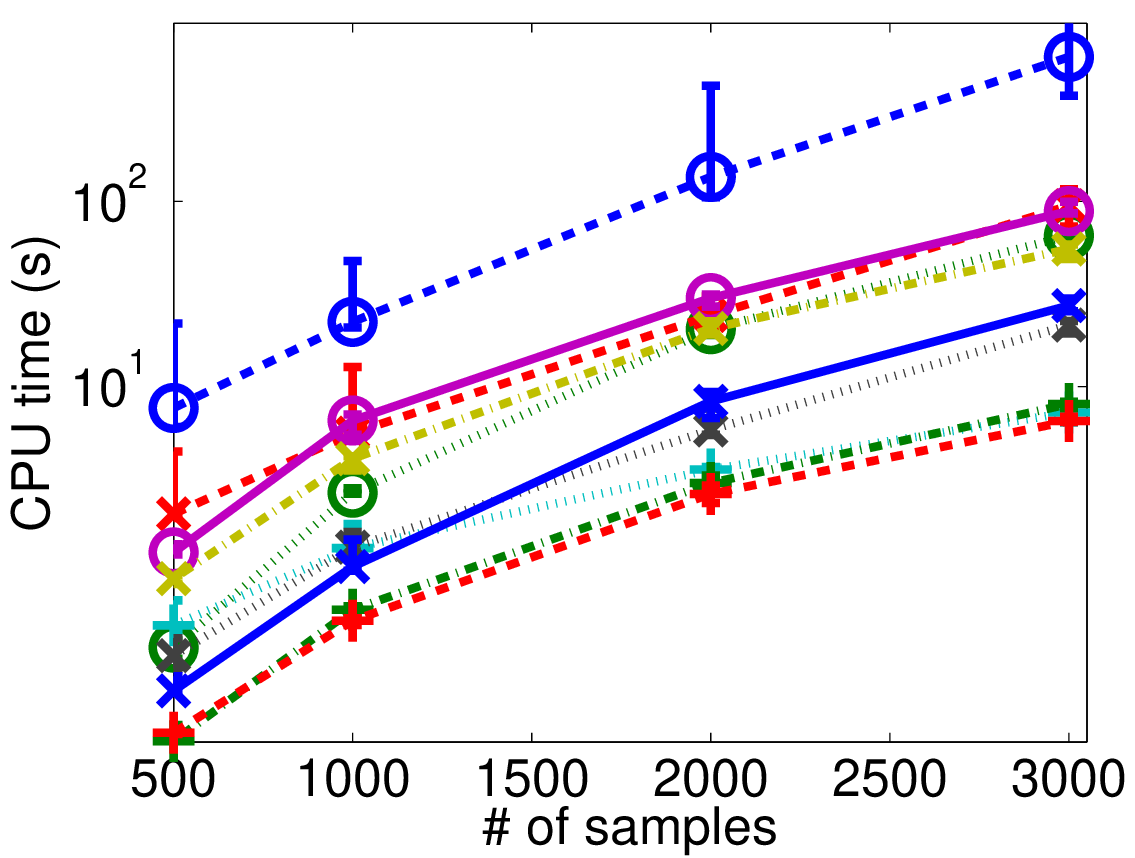}
}
\subfigure[Waveform]{
\label{fig:SampleScalingWaveform}
\includegraphics[height=4.7cm,clip]{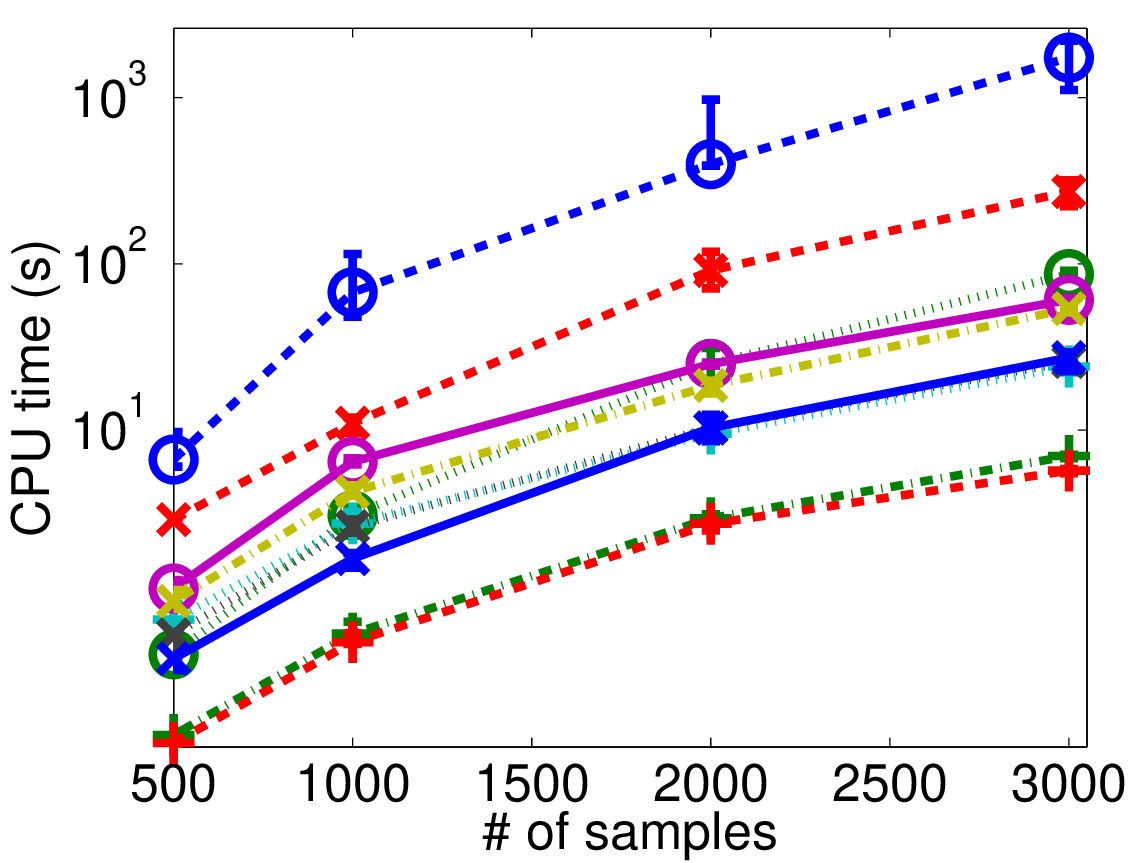}
}
\\
\vspace{0.2cm}
\subfigure{ 
\includegraphics[height=1.5cm,clip]{legends.eps}
}
\caption{CPU time as a function of the sample size}
\label{fig:SampleScaling}
\end{center}
\vspace{-5mm}
\end{figure*}

In Figure~\ref{fig:SampleScaling} the number of training samples is increased from 500 to 3000. 
The number of kernels is fixed to 20.
SILP shows fairly good computational efficiency among the methods with block 1-norm regularization. 
Comparing the result of Figure~\ref{fig:NumKernelScaling},
while SILP is not fast for a large number of kernels, SILP converges relatively fast for a small number of kernels and scales well against the number of samples.
The CPU time of SpicyMKL and ElastMKL
is comparable to that of other methods. 
In particular, ElastMKL shows the best performance in Splice, Twonorm and Waveform.


\section{Conclusion and future direction}
\label{sec:conclusion}
In this article, we have proposed a new efficient training algorithm for MKL with general convex loss functions and 
general regularizations.
The proposed SpicyMKL algorithm generates a sequence of primal variables by
iteratively optimizing a sequence of smooth minimization problems.
The outer loop of SpicyMKL is a proximal minimization method and it converges
super-linearly. The inner minimization is efficiently carried out by the Newton method. 
We introduced a generalized block norm formulation of MKL regularization that includes non-sparse regularizations such as Elastic-net MKL and block $q$-norm MKL,
and derived the connections between our block norm formulation and the kernel weights. 
Then we derived a general optimization method that is applicable to the general block norm framework.
We also gave an efficient one step optimization method for regularizations with smooth dual (e.g., elastic-net and block $q$-norm).

We have shown through numerical experiments that SpicyMKL scales well with increasing number of kernels
and it has similar scaling behavior against the number of samples to conventional methods. 
It is worth noting that SpicyMKL does not rely on solving LP or QP problems,
which itself can be a challenging task; accordingly SpicyMKL can reliably
obtain a precise solution.
SpicyMKL with the logistic loss and elastic-net regularization has shown the best performance while 
showing fairly good accuracies.

Future work includes a second order modification of the update rule of the primal variables, and 
combination of the techniques developed in SpicyMKL and
wrapper methods
\citep{JMLR:Sonnenburg+etal:2006,JMLR:Rakotomamonjy+etal:2008,NIPSWS:Chapelle:2008}.

It would also be interesting to 
develop an efficient decomposition method to deal with a large sample size problem.
Since the dual problem \eqref{eq:alfunc} of the inner loop is an $N$ dimensional optimization problem,
when the number of samples  $N$  is large, the naive Newton method or other gradient descent type methods might be hard to be applied.
In such a situation, the block-coordinate descent algorithm might be useful.
That is, decompose $N$ variables $\{\rho_i\}_{i=1}^N$ into some groups (say $B$ groups $\{\rho_{I_i}\}_{i=1}^B$) and iteratively minimize the objective function 
with respect to one group with other groups fixed:
$$
\rho_{I_i}^{k+1} \leftarrow \mathop{\arg\min}_{\rho_{I_i}} 
\varphi_{\gammam{(t)}}((\rho_{I_1}^{k+1},\dots,\rho_{I_{i-1}}^{k+1},\rho_{I_i},\rho_{I_{i+1}}^{k},\dots,\rho_{I_{B}}^{k}); \alpha^{(t)},b^{(t)}).
$$
(see Section 5.4.3 of \citet{Book:Znagwill:1969} and Proposition 2.7.1 of \citet{Book:Bertsekas:1999}).
In a similar sense, Sequential Minimal Optimization (SMO) algorithm
\citep{NIPS:Platt:99} might be useful for some loss function classes such
as the hinge loss.
We leave these important issues for future work.


\section*{Acknowledgement}
We would like to thank anonymous reviewers for their constructive
comments, which improved the quality of this paper. We would also like
to thank Manik Varma, Marius Kloft, Alexander Zien, and
Cheng Soon Ong for helpful discussions.
This work was partially supported by MEXT KAKENHI 22700289 and 22700138. 
\appendix

\section{Proof of Theorem \ref{th:superlinear}}
\label{sec:ProofTheorem}
Combining Eqs.~\eqref{eq:derivedual1}, \eqref{eq:derivedual2delta}, and
\eqref{eq:derivedual3}, we can rewrite the dual of the proximal MKL
problem~\eqref{eq:prox_min} as follows:
\begin{align*}
\minimize_{\begin{subarray}{c}
\rho\in\Real^{N}\\
u\in\Real^{MN}
\end{subarray}}\quad
\biggl\{
\textstyle
&
L^\ast(-\rho)+
\sum\limits_{m=1}^M
\delta_{C}^{(m)}(u_m) \notag \\
&+
\sum\limits_{m=1}^M
\bigl( {\alpha_m^{(t)}}\T K_m(\rho-u_m)+\frac{\gammam{(t)}}{2}\|\rho-u_m\|_{K_m}^2 \bigr)
+
\!
b^{(t)}
\boldone^\top \rho \!-\!
\frac{\gammab{(t)}}{2} \!\!\bigl(\textstyle \boldone^\top \rho  \bigr)^2\! \biggr\},
\end{align*}
where the maximization is turned into minimization and we redefined
$u_m/\gamma^{(t)}$ as $u_m$.
This formulation is known as the {\it augmented Lagrangian} 
for the dual of the MKL optimization problem~\eqref{eq:MKL_primal},
which can be expressed as follows:
\begin{align*}
 \minimize_{
\begin{subarray}{c}
\rho\in\Real^{N},\\
u\in\Real^{MN}
\end{subarray}}\quad&L^{\ast}(-\rho)+\sum_{m=1}^{M}\delta_C^{(m)}(u_m),\\
\subjectto\quad & K_m^{1/2}(\rho-u_m)=0\quad(m=1,\ldots,M),\\
&\mathbf{1}\T\rho=0.
\end{align*}
 $K_m^{1/2} \alpha_m^{(t)}$ and $b^{(t)}$ are the Lagrangian multipliers
 corresponding to the above $M+1$ equality constraints~\citep{JOTA:Hestenes:1969,Opt:Powell:1969,ISPL:Tomioka+Sugiyama:2009}.
We apply Proposition 5.11 of \citet{Book:Bertsekas:1982}
so that we obtain 
\[
\sum_{m=1}^M \|\alpha_m^{(t)} - \alpha_m^{\ast} \|_{K_m}^2 + (b^{(t)} - b^{\ast})^2 
\to 0~~~~(\text{as}~t\to \infty). 
\]
Thus there is sufficiently large $T$ such that for all $t \geq T$, 
$(\alpha^{(t)},b^{(t)})$ is inside the $\delta$-neighborhood of $(\alpha^{\ast},b^{\ast})$.
Therefore we can assume that 
\Eqref{eq:LocalStrongConv} holds at $(\alpha^{(t)},b^{(t)})$ for all $t \geq T$.
Finally Theorem 3 of \citet{JMLR:Tomioka+Suzuki+Sugiyama:2011} (and its proof) gives the assertion
(see also the proof of Proposition 5.22 of \citet{Book:Bertsekas:1982}
where $\phi(t) = t^2$ and $\nabla \phi(t) = t$ are substituted for our setting).


\end{document}